\documentclass{fairmeta}

\usepackage{latexsym}
\usepackage[T1]{fontenc}
\usepackage[utf8]{inputenc}
\usepackage{microtype}
\usepackage{graphicx}
\usepackage{booktabs}
\usepackage{algorithm}
\usepackage{algpseudocode}
\usepackage{tabularx}
\usepackage{pifont}
\usepackage[nointegrals]{wasysym}
\usepackage{xcolor}
\usepackage{placeins}
\usepackage{pgfplots}
\usepackage{xurl} 
\pgfplotsset{compat=1.18}
\definecolor{swgreen}{RGB}{40,167,69}
\definecolor{swred}{RGB}{220,53,69}
\definecolor{swyellow}{RGB}{255,193,7}
\newcommand{\swyes}{{\color{swgreen}\ding{51}}}
\newcommand{\swno}{{\color{swred}\ding{55}}}
\newcommand{\swhalf}{{\color{swyellow}\LEFTcircle}}
\newcolumntype{Y}{>{\centering\arraybackslash\hspace{0pt}}X}

\IfFileExists{assets/optimistic.tfm}{}{%
  \expandafter\def\csname Inter@scale\endcsname{0.88}}
\IfFileExists{assets/logometa.pdf}{}{%
  \patchcmd{\mymaketitle}{\includegraphics[width=1.5cm]{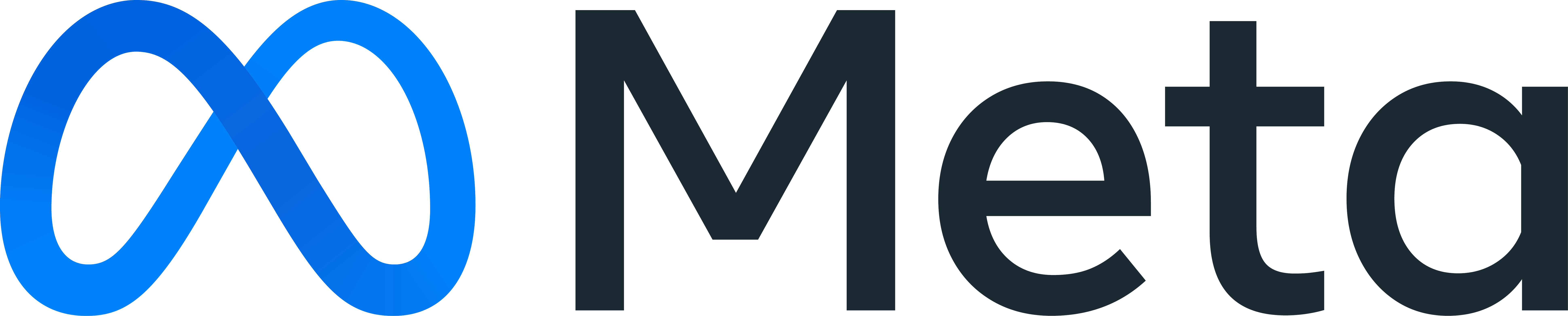}}{}{}{}}

\renewcommand{\cite}{\citep}
\title{\textsc{SurveyWeave}: An Agentic Framework for Automated Survey Generation}

\author[1]{Franklin Zhang}
\author[2]{Xiao Yang}
\author[2]{Xin Luna Dong}
\author[3]{Juhua Hu}
\affiliation[1]{University of Washington}
\affiliation[2]{Meta Reality Labs}
\affiliation[3]{University of Washington Tacoma}

\abstract{
Surveys help researchers keep up with fast-moving fields, but they take weeks to months to write. Automated survey generation (AutoSurvey) can alleviate this situation, but it is challenging to draw on hundreds to thousands of precisely scoped papers, choose among many valid structures, and stay faithful to its sources while offering insights. We present \textsc{SurveyWeave}, an agentic architecture that decomposes the problem into four tasks coordinated by three agents, each reviewing others' work and conducting revisions based on feedback. We also introduce \textsc{WeaveEval}, a five-dimension evaluation grounded in prior surveys. A comprehensive comparison shows that \textsc{SurveyWeave} outperforms Codex by 8 points and the strongest AutoSurvey baseline by 6 points.

}

\date{\today}
\correspondence{Franklin Zhang at \email{frankyz@uw.edu}}

\begin{document}
\maketitle

\section{Introduction}

\begin{figure*}[t]
  \centering
  \begin{minipage}[c]{0.65\textwidth}
    \centering
    \includegraphics[width=\linewidth]{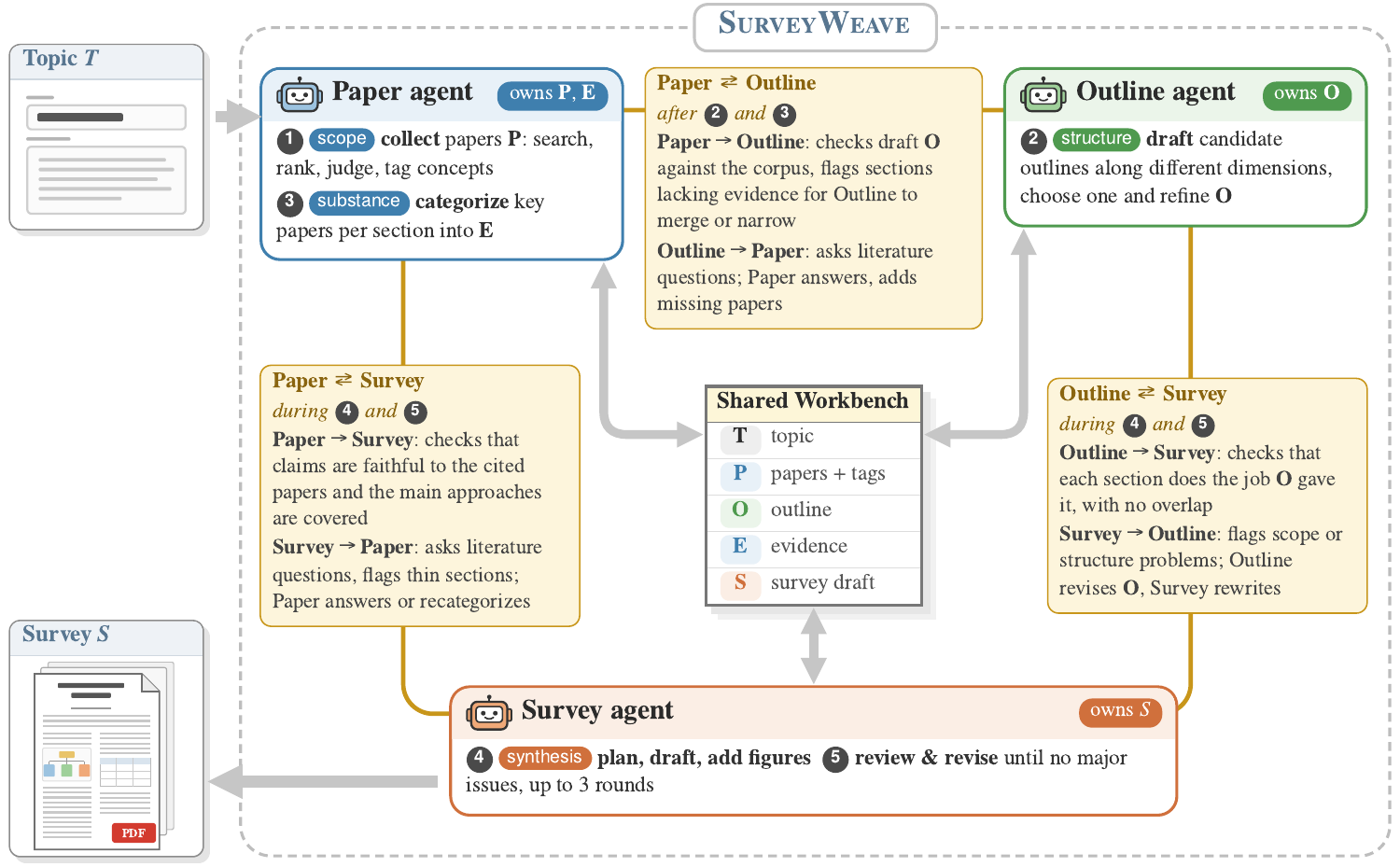}\\[1pt]
    {\small (a) \textsc{SurveyWeave}}
  \end{minipage}\hfill
  \begin{minipage}[c]{0.33\textwidth}
    \centering
    \includegraphics[width=\linewidth]{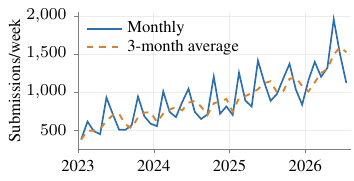}\\[1pt]
    {\small (b) Weekly arXiv submissions}\\[6pt]
    \includegraphics[width=\linewidth]{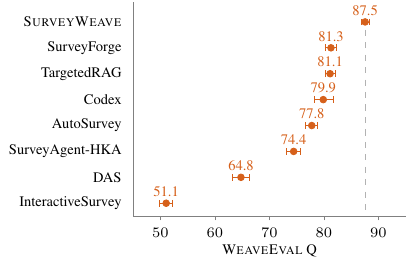}\\[1pt]
    {\small (c) Headline results}
  \end{minipage}
  \caption{(a) Overview of the \textsc{SurveyWeave} agentic architecture. (b) Weekly arXiv submissions in cs.CL, cs.LG, cs.AI, and cs.IR. (c) \textsc{SurveyWeave} outperforms a state-of-the-art general-purpose coding agent (Codex) and AutoSurvey systems on \textsc{WeaveEval} (mean Q with 95\% CI over topics; Table~\ref{tab:main}).}
  \label{fig:crownjewel}
\end{figure*}

As AI facilitates research exploration, there have been growing scientific discoveries in computer science. As shown in Figure~\ref{fig:crownjewel}(b), the number of uploaded papers in four categories of arXiv (cs.CL, cs.LG, cs.AI, and cs.IR) has grown from an average of 580 per week in Jan--Jul 2023 to 1,377 per week in Jan--Jul 2026 ($2.4\times$). Surveys are one of the most effective ways for researchers to enter a new area and to keep up with a fast-moving one. Writing a good survey, however, is becoming increasingly expensive with the fast growth of papers. Authors must find and read hundreds to thousands of papers, and the result goes out of date quickly. Large language models (LLMs) can read and summarize texts at a scale no human can match, and they write fluent long-form text. This makes automated survey generation (AutoSurvey) both appealing and increasingly feasible~\cite{wang2024autosurvey,zheng2025automation}.

AutoSurvey appears to be a special case of the \textit{deep research} problem~\cite{huang2025deepresearchagents,xu2025deepresearchsurvey}: an agent searches for sources, reads them, and writes a grounded report. However, AutoSurvey raises challenges well beyond typical deep research. First, the evidence is at a much larger \textit{scale} and needs precise \textit{boundaries}. A survey draws on hundreds to thousands of papers rather than dozens of web pages, yet it must exclude neighboring topics with overlapping vocabulary (e.g., modularized RAG for an agentic RAG survey). Second, the \textit{structure} is far more complex. Sub-areas overlap and can be organized in many valid ways, and the chosen organization must read as an easy-to-understand linear narrative in which later sections build on earlier ones. Third, scientific surveys are held to a high standard of {\em quality} and \textit{insight}. They must stay coherent across tens of pages and remain faithful to the cited papers. Beyond summarizing, a good survey compares approaches, traces how ideas evolved, and identifies open problems.

Systems built specifically for AutoSurvey are often still rough and may fall short at every task, such as collecting irrelevant or low-quality papers into the corpus, producing a sub-optimal outline, or drafting overlapping subsections. Most importantly, they forget that ``writing is rewriting'': they either follow a fixed plan--retrieve--write pipeline~\cite{wang2024autosurvey,liang2025surveyx,yan2025surveyforge,wang2025mapreducev2}, or use an agentic system that only partially revisits previous decisions or writing~\cite{itersurvey,mapreducev3,deepsurvey,surveygeni,fiksurvey,das,scisage,surveyagenthka}.

Our solution, \textsc{SurveyWeave}, rests on two ideas. First, we decompose AutoSurvey into four tasks, each deepening the system's understanding of the area: \textit{paper collection} establishes its \textit{scope}, \textit{outline generation} defines its \textit{structure}, \textit{paper categorization} indicates its \textit{substance}, and \textit{survey writing} details its \textit{synthesis}. Within each task, we carefully design algorithms to generate high-quality intermediate results. 
For example, we batch LLM judgment of papers guided by a distilled ranker to filter out papers from neighboring fields, we generate alternative outlines that each organize the field along a different dimension and compare them before final selection, and we make a trial draft that exposes overlap between sections before writing the full survey. Second, rather than running these tasks as a one-pass pipeline, we coordinate them with an \textit{agentic system} that revisits earlier decisions, as human authors do. Prior systems assign agents to tasks and typically rely on a single reviewer. \textsc{SurveyWeave} instead assigns each of three agents to an \textit{entity} it owns (the papers, the outline, or the survey), so each agent keeps the context it builds up across tasks. Each agent asks the others to review its output from their own perspectives and sends revision requests to owner agents (Figure~\ref{fig:crownjewel}(a)).

Evaluating surveys is in itself difficult. Existing benchmarks either judge a survey without reference to human-written surveys, which makes coverage and novelty hard to assess~\cite{bloom}, or compare it against a single human survey, which rewards similarity to that survey rather than quality~\cite{das}. We evaluate \textsc{SurveyWeave} with \textsc{WeaveEval}, which combines reference-free and reference-grounded judgments across five dimensions: evidence reliability, coverage, synthesis, new insight, and presentation. 

In summary, we make three contributions and compare with earlier systems in Table~\ref{tab:related-features}. We will release our code and data upon acceptance.

\noindent 1) We present \textsc{SurveyWeave}, an agentic architecture for automated survey generation. Agents each own an entity and review one another. To our knowledge, it is the first architecture in which problems found during writing can revise the corpus, the outline, and the paper categorization.

\noindent 2) The \textsc{SurveyWeave} agents collaborate on paper collection, outline generation, paper categorization, and survey writing. We develop new techniques (e.g., distilled-ranker-guided LLM filtering, comparison of alternative outlines, and trial drafting) to generate high-quality results.

\noindent
3) We propose the \textsc{WeaveEval} evaluation framework and evaluate \textsc{SurveyWeave} on 36 topics against 7 baselines (\S\ref{sec:setup}). \textsc{SurveyWeave} scores highest by 8 points over Codex and 6 points over the strongest AutoSurvey baseline (Figure~\ref{fig:crownjewel}(c)). 

\section{Related Work}
\label{sec:related}

\begin{table*}[t]
\centering
\footnotesize
\setlength{\tabcolsep}{2pt}
\renewcommand{\arraystretch}{1.1}
\caption{Comparison on advanced features; systems ordered by publication date within groups. \swyes: supported; \swno: not supported; \swhalf: partially supported (not in a principled or comprehensive way). {\sc SurveyWeave} is the only system that supports all of the following: applies LLM-based filtering and collects background papers in paper collection; derives outlines from a taxonomy, uses web and prior surveys as external evidence, and generates multiple outlines for mixing / selection; considers paper quality; generates figures and tables; adopts an agentic architecture, uses entity-based agents, conducts cross-agent review and backward revision.}
\label{tab:related-features}
\begin{tabularx}{\textwidth}{l *{11}{Y}}
\toprule
& \multicolumn{2}{c}{\textbf{Collection}}
& \multicolumn{3}{c}{\textbf{Outline}}
& \textbf{Cate.}
& \textbf{Surv.}
& \multicolumn{4}{c}{\textbf{Architecture}} \\
\cmidrule(lr){2-3}\cmidrule(lr){4-6}\cmidrule(lr){7-7}\cmidrule(lr){8-8}\cmidrule(lr){9-12}
\textbf{System}
& LLM filtering
& Bkg papers
& Taxo-nomy based
& Exter. evid.
& Multi outline
& Quality based
& Fig/ Table gen
& Agentic
& Ent. based agents
& Cross agent rev.
& Back revision \\
\midrule
\multicolumn{12}{l}{\bf\em Pipelined approaches} \\
AutoSurvey~\cite{wang2024autosurvey}         & \swno   & \swno  & \swno   & \swno   & \swhalf & \swno  & \swno   & \swno   & \swno  & \swno   & \swno   \\
SurveyX~\cite{liang2025surveyx}              & \swyes  & \swno  & \swhalf & \swhalf & \swno   & \swno  & \swyes  & \swno   & \swno  & \swno   & \swhalf \\
SurveyForge~\cite{yan2025surveyforge}        & \swno   & \swno  & \swno   & \swhalf & \swhalf & \swyes & \swno   & \swhalf & \swno  & \swno   & \swno   \\
LLM$\times$MapReduce-V2~\cite{wang2025mapreducev2} & \swno & \swno & \swhalf & \swno & \swyes  & \swno  & \swno   & \swno   & \swno  & \swhalf & \swno   \\
\midrule
\multicolumn{12}{l}{\bf\em Agentic approaches} \\
SciSage~\cite{scisage}                       & \swno   & \swno  & \swno   & \swno   & \swhalf & \swyes & \swhalf & \swyes  & \swno  & \swhalf & \swhalf \\
SurveyGen-I~\cite{surveygeni}                & \swyes  & \swyes & \swno   & \swhalf & \swno   & \swno  & \swyes  & \swhalf & \swno  & \swno   & \swno   \\
LLM$\times$MapReduce-V3~\cite{mapreducev3}   & \swno   & \swno  & \swyes  & \swno   & \swno   & \swno  & \swyes  & \swyes  & \swno  & \swhalf & \swno   \\
IterSurvey~\cite{itersurvey}                 & \swno   & \swyes & \swno   & \swno   & \swno   & \swno  & \swyes  & \swhalf & \swno  & \swno   & \swno   \\
DeepSurvey~\cite{deepsurvey}                 & \swyes  & \swyes & \swyes  & \swno   & \swno   & \swno  & \swhalf & \swyes  & \swno  & \swno   & \swno   \\
FIKSurvey~\cite{fiksurvey}                   & \swno   & \swno  & \swno   & \swno   & \swno   & \swno  & \swyes  & \swhalf & \swno  & \swno   & \swhalf \\
DAS~\cite{das}                               & \swyes  & \swno  & \swyes  & \swno   & \swno   & \swno  & \swyes  & \swyes  & \swno  & \swno   & \swhalf \\
SurveyAgent-HKA~\cite{surveyagenthka}        & \swyes  & \swyes & \swyes  & \swhalf & \swno   & \swyes & \swno   & \swyes  & \swno  & \swno   & \swhalf \\
\midrule
\textbf{\textsc{SurveyWeave}}                & \swyes  & \swyes & \swyes  & \swyes  & \swyes  & \swyes & \swyes  & \swyes  & \swyes & \swyes  & \swyes  \\
\bottomrule
\end{tabularx}
\end{table*}

\paragraph{Deep research.}
Deep research agents answer open-ended questions by repeatedly planning searches, reading sources, and writing a report grounded in what they find~\cite{huang2025deepresearchagents,xu2025deepresearchsurvey}. Commercial systems such as OpenAI Deep Research~\cite{openai2025deepresearch}, Gemini Deep Research~\cite{google2024deepresearch}, and Perplexity Deep Research~\cite{perplexity2025deepresearch}, as well as research prototypes such as STORM~\cite{shao2024storm} and Co-STORM~\cite{jiang2024costorm}, show that LLM agents can write long, cited reports on unfamiliar topics. STORM, for example, writes Wikipedia-style articles by simulating conversations between a topic expert and writers with different perspectives. These systems, however, are designed for reports grounded in dozens of web pages. They do not collect papers at survey scale within precise topic boundaries, organize them into a taxonomy whose sections build on one another, or identify the work that defines each sub-area.

\paragraph{Automated survey generation.}
Early systems built specifically for automated survey generation, such as AutoSurvey~\cite{wang2024autosurvey}, SurveyForge~\cite{yan2025surveyforge}, SurveyX~\cite{liang2025surveyx}, and LLM$\times$MapReduce-V2~\cite{wang2025mapreducev2}, follow a largely fixed retrieve--outline--write pipeline. They retrieve papers mainly by embedding similarity, generate an outline in one or a few LLM passes, and write each section from the papers retrieved for it. These systems established the task and its evaluation protocols~\cite{wang2024autosurvey,yan2025surveyforge}. However, each stage runs only once, so errors in retrieval or organization carry through to the final survey. 

More recent systems, summarized in Table~\ref{tab:related-features}, add iteration and feedback. We group them by how far back that feedback can reach. The first group builds the outline iteratively from evidence. IterSurvey~\cite{itersurvey} alternates between retrieval and outline updates; LLM$\times$MapReduce-V3~\cite{mapreducev3} merges critiques of the outline from individual papers, guided by a human; DeepSurvey~\cite{deepsurvey} refines the outline against paper analyses. In all three, the outline is final before writing begins. The second group reviews and repairs the draft but keeps the corpus and outline fixed. FIKSurvey~\cite{fiksurvey} scores the outline, content, and citations with a helper LLM. DeepSurvey~\cite{deepsurvey} also verifies each citation against its source. DAS~\cite{das} explicitly reactivates only writing states, so review cannot alter upstream decisions. The third group lets feedback reach earlier tasks, but only partially. SurveyGen-I~\cite{surveygeni} revises its outline during writing, but only for subsections that have not yet been written. The reviewers in SciSage~\cite{scisage} and SurveyAgent-HKA~\cite{surveyagenthka} can trigger additional retrieval, but never regenerate the outline or reassign papers to sections.

Unlike these systems, which assign agents to tasks and usually rely on a single reviewer, \textsc{SurveyWeave} assigns each agent to an entity it owns: the papers, the outline, or the survey. The agents review one another's work, which helps revise the corpus, the outline, and the paper categorization during survey writing. 

\section{\textsc{SurveyWeave} Agentic Framework}
\label{sec:surveyweave}

Upon receiving a research topic $T$, which includes a topic name and a description, \textsc{SurveyWeave} produces a publication-quality survey $S$, in the form of a PDF document compiled from \LaTeX{} source.


\subsection{Agentic system}
\label{subsec:overview}
We first define four critical tasks in survey writing, mimicking how a human researcher progressively understands a domain and writes a survey.

\subsubsection{Tasks in writing a survey}
\paragraph{Paper collection.}
Paper collection takes as input the research topic $T$, and outputs a set of papers ${\bf P} = \{(p, d_p, \bar{C}_p)\}$ that are relevant;
here, each paper $p$ is annotated with a description $d_p$ and a set $\bar{C}_p$ of \textit{technical concept tags}, describing technical topics covered by the paper in text phrases. This task establishes the \textit{scope} of the area: which work belongs to the topic.

\paragraph{Outline generation.}
Outline generation takes as input the topic $T$ and the collected paper corpus $\bf P$, and outputs a hierarchical outline ${\bf O}=({\bf V}, {\bf R})$. We denote each section or subsection by a node $v \in \bf V$, 
and the subsection relationships by $\bf R$.  Building on the scope, this task reveals the \textit{structure} of the area: how its sub-areas divide the topic and how they depend on one another.

\paragraph{Paper categorization.}
Paper categorization takes as input the paper corpus $\bf P$ and the outline $\bf O$, and assigns each outline node $v \in \bf V$ an evidence set $\bar{E}_v \subseteq {\bf P}$ of papers that can be used to write that section. We denote all evidence sets by ${\bf E}$. This task fills the structure with \textit{substance} by identifying the papers that define each sub-area and its main lines of progress.

\paragraph{Survey writing.}
Survey writing takes as input the outline and its evidence sets $({\bf O}, {\bf E})$, and outputs a full survey $S$. This task turns structured evidence into a \textit{synthesis}: it explains how ideas relate within and across sub-areas and what the field has collectively achieved.

\subsubsection{Agentic architecture}
\label{sec:agentic}
The simplest way to conduct the aforementioned four tasks is a pipeline that runs them in order: paper collection gathers the topic corpus $\bf P$, outline generation creates the outline $\bf O$, paper categorization assigns each outline node its evidence set, yielding $\bf E$, and survey writing produces the final survey $S$. Each task consumes the output of the previous ones, and earlier outputs are never revisited.

The pipeline seems to follow how researchers produce a survey, but cannot reflect the iterative process in writing. As researchers write, they often find weaknesses in their taxonomy and outline, and go back to revise them. A pipeline, by contrast, allows early mistakes and misguided intuitions to become permanently built in and limit later decisions. Our intuition is therefore to design an agentic system where agents review each other's work and revisit earlier decisions. As we design the agentic system, we follow three principles.

\paragraph{Agents.} First, context from early actions can inform later steps: an agent that collected the papers is best placed to judge which of them a section needs. Thus, instead of creating an agent for each task, we \textit{create agents each associated with a particular entity}: the papers, the outline, and the survey. Each agent takes on defined responsibilities centered on the entity it represents. The \textbf{Paper agent} owns the corpus $\bf P$. It performs paper collection (\S\ref{sec:impl-collection}) and paper categorization (\S\ref{sec:impl-categorization}). The \textbf{Outline agent} owns the outline $\bf O$. It performs outline generation (\S\ref{sec:impl-outline}). The \textbf{Survey agent} owns the survey $S$. It performs survey writing (\S\ref{sec:impl-writing}), including planning, drafting, figure generation, and revision, and decides when $S$ is finished. An agent may hand a self-contained part of its task, such as drafting one section or one figure, to a \textit{sub-agent}: a short-lived LLM instance that receives only the inputs for that part and returns its result to the agent. Sub-agents own no entity and take no part in check-in points, so \textsc{SurveyWeave} has three agents.

\paragraph{Memory.} Second, the agents' tasks depend on each other; thus, we maintain \textit{a common \textit{workbench} (\textit{memory}) that stores their intermediate results, and an agent hands work to another once the data it needs is in the workbench}. Specifically, the workbench contains $(T, {\bf P}, {\bf O}, {\bf E}, S)$ and a shared review ledger. In the forward pass, the Paper agent fills in $\bf P$ and hands off to the Outline agent, which fills in $\bf O$ and hands back; the Paper agent then fills in $\bf E$ and hands off to the Survey agent, which generates $S$. The check-in points below send requests back to earlier owners, and reviews are recorded in the ledger.

\paragraph{Check-in points.} Third, LLM agents tend to focus on their assigned tasks and fully trust their inputs, rarely reviewing outputs from earlier tasks. For this reason, we \textit{build review and revision into our agent workflow as explicit steps}: each agent may trigger other agents to review its output from their perspectives, and can also pass the identified problems to other agents for fixes. The check-in points are listed below; the final review of the survey repeats while major issues remain, for up to three rounds.

\noindent 1) \textbf{Paper \& Outline agents:} The Outline agent sends its draft outline to the Paper agent, which checks it against the corpus and answers the Outline agent's questions about the literature, adding papers the corpus lacks. After categorization, the Paper agent returns sections that still lack evidence, and the Outline agent restructures them, for example by merging or narrowing them.

\noindent 2) \textbf{Paper \& Survey agents:} The Paper agent checks that the survey's claims are faithful to the cited papers and that it covers the main approaches in the corpus. In turn, the Survey agent sends it questions about the literature and sections that lack evidence, which the Paper agent answers, re-categorizes, or resolves by collecting new papers.

\noindent 3) \textbf{Outline \& Survey agents:} The Outline agent checks that the survey follows the outline: each section does the job the outline gave it, and sections do not overlap. When writing exposes a scope or structure problem, the Survey agent asks the Outline agent to revise the outline; the Paper agent then re-categorizes the changed sections, and the Survey agent rewrites them.

Figure~\ref{fig:crownjewel} depicts the agentic architecture and Appendix~\ref{app:algorithm} details the algorithm.

\subsection{Task Implementation}
\label{sec:component-impl}
Having introduced the four tasks involved in generating a survey and the agentic architecture that coordinates them, we now examine each task, non-trivial on its own. 

\subsubsection{Paper Collection}
\label{sec:impl-collection}
Paper collection takes a research topic $T$ as input, queries an external paper corpus (e.g., arXiv), and outputs a set of papers ${\bf P} = \{(p, d_p, \bar{C}_p)\}$.

We consider three types of papers: (1) the \emph{core} set ${\bf P}_{\mathrm{core}}$, which contains papers that directly and explicitly contribute to $T$; (2) the \emph{background} set ${\bf P}_{\mathrm{bg}}$, which contains papers that do not necessarily address problems in $T$ but provide important background, such as a canonical LLM paper in a RAG survey; and (3) the \emph{survey} set ${\bf P}_{\mathrm{surv}}$, which contains prior surveys on the same or related topics. ${\bf P}={\bf P}_{\mathrm{core}} \cup {\bf P}_{\mathrm{bg}} \cup {\bf P}_{\mathrm{surv}}$.


At first glance, paper collection looks like a simple keyword-search problem. However, it poses two challenges. First, it requires much higher precision than typical search: including papers from a neighboring topic can pollute the survey, for example by collecting many text-RAG papers for an MM-RAG survey. Second, we must choose ${\bf P}_{\mathrm{bg}}$ carefully, so that the survey is grounded in essential background knowledge from authoritative papers without going into unnecessary detail. 

\paragraph{Step 1: Search for core-paper candidates.} We first generate a candidate set $\hat{{\bf P}}_{\mathrm{core}}$ that is intended to cover all core papers, that is, ${\bf P}_{\mathrm{core}} \subseteq \hat{{\bf P}}_{\mathrm{core}}$. To do so, we prompt an LLM to generate multiple keyword queries and run both keyword and semantic (embedding-based) search. The resulting candidate set can be noisy, but a focused survey needs a clean core set (Challenge~\#1). We therefore rank and filter the candidates in the next two steps.

\paragraph{Step 2: Rank by relevance.} We rank the papers in $\hat{{\bf P}}_{\mathrm{core}}$ by their relevance to $T$. Search similarity scores often miss subtle distinctions between scientific terms (e.g., RAG vs.\ MM-RAG). LLM ratings capture these distinctions but are too expensive to generate for the thousands to tens of thousands of candidates that are typical. We therefore distill logistic regression models per input topic from LLM-generated labels on a small set ($\sim$500) of papers and use these low-cost models to score and rank all candidates.

\paragraph{Step 3: Judge and label with an LLM.} Given the ranked list, we next choose a cutoff to finalize ${\bf P}_{\mathrm{core}}$. To save cost and latency, we send papers to an LLM in batches, in ranked order (batch size 100). For each paper $p$, the LLM (1) judges whether $p$ is relevant to $T$, and (2) generates the description $d_p$ and the concept tags $\bar{C}_p$. We stop when two consecutive batches contain fewer than 4\% relevant papers. The papers judged relevant form ${\bf P}_{\mathrm{core}}$.

\paragraph{Step 4: Collect background papers.} Our intuition is that background papers are frequently cited by papers in ${\bf P}_{\mathrm{core}}$ and are \textit{influential}, as reflected in their overall citation counts (Challenge~\#2). We therefore follow citations one hop away from ${\bf P}_{\mathrm{core}}$ and rank the cited papers by how often they are cited within ${\bf P}_{\mathrm{core}}$, breaking ties by overall citation count. We select papers up to a budget of $B = \lceil \alpha \cdot \sqrt{|{\bf P}_{\mathrm{core}}|}\,\rceil$ (we set $\alpha=16$). The square-root budget reflects our intuition that a relatively small body of influential work is enough to supply the necessary context.

\subsubsection{Outline Generation}
\label{sec:impl-outline}
Outline generation takes the collected papers ${\bf P}$ as input and generates the outline ${\bf O} = ({\bf V}, {\bf R})$. The outline largely determines how understandable a survey is: it must present a clear big picture, and it must also form a coherent narrative in which later sections build on earlier ones.

Outline generation can be associated with taxonomy generation: we identify the sub-areas of the topic, and the outline follows naturally from the resulting taxonomy. The challenges come from how taxonomies and outlines differ. First, sub-areas of a research topic often overlap, so the same sub-areas can be organized into a taxonomy in several valid ways, and some of these taxonomies are much easier to understand than others. Second, a taxonomy imposes no order on sibling nodes, whereas an outline is essentially linear, and the order of sections affects how easily readers can follow the survey. 

\paragraph{Step 1: Cluster concepts.} This step generates the nodes of the taxonomy from the concept tags assigned to core papers during paper collection, $\{\bar{C}_p : p \in {\bf P}_{\mathrm{core}}\}$. Because these concepts often overlap, we group them using \textit{agglomerative clustering}~\cite{sorensen1948method} into at most $K=200$ clusters, a size that keeps nearly all important concepts while remaining small enough for the LLM to reason over without noise. Each resulting cluster $G_i$ ($1 \le i \le K$) is named after its most frequent tag and is weighted by the number of core papers it covers
$
  N_i = \left|\left\{p \in {\bf P}_{\mathrm{core}} : \bar{C}_p \cap G_i \neq \emptyset\right\}\right|.
$

\paragraph{Step 2: Collect outline evidence.} To decide how best to organize the concept clusters (Challenge~\#1), we draw on existing knowledge of the research topic from two sources: (1) the Web and (2) the prior surveys in ${\bf P}_{\mathrm{surv}}$. From each source, we prompt the LLM to generate a report that describes the topic's central concepts, its key mechanisms, and the boundaries between sub-areas that remain contested. These reports serve as evidence for the next step.

\paragraph{Step 3: Construct the outline.} We prompt an LLM to generate a survey outline from the concept clusters produced in Step~1 and the evidence reports collected in Step~2. Because the key concepts of a field can be organized in several ways (Challenge~\#1), we generate multiple candidate outlines. Each candidate organizes the field along one \textit{dimension}, such as methodology or the successive generations of techniques. The LLM then evaluates the candidates on overall structure, balance of section sizes (measured by number of papers), and section ordering for understandability (Challenge~\#2). 
It compares their strengths and weaknesses, selects one organizing dimension, and incorporates useful elements from the others. 

\subsubsection{Paper Categorization}
\label{sec:impl-categorization}
Paper categorization takes the paper collection ${\bf P}$ and the outline ${\bf O} = ({\bf V}, {\bf R})$ as input; for each subsection $v \in {\bf V}$, it identifies a set of key papers $\bar{E}_v$, and collectively outputs the evidence sets ${\bf E}$.

Paper categorization is essentially a classification problem, and we can use the concept tags assigned during paper collection to solve it. The main challenge is \textit{scale}. A single section may relate to hundreds or even thousands of papers, yet most of them make only marginal contributions. 
We thus rank each section's candidate papers by quality: for earlier papers, we measure quality by citation count; for recent papers whose citation counts are not yet informative, we use a quality score that the LLM generates together with the paper's summary.

\subsubsection{Survey Writing}
\label{sec:impl-writing}
Survey writing takes the outline and its evidence sets $({\bf O}, {\bf E})$ as input and produces the final survey $S$. This appears to be simply a summarization and writing problem, which plays to the strengths of LLMs. However, 
generating a high-quality, comprehensive survey from hundreds of papers in a single pass is hard because the context window is limited and the domain is complex. 
We therefore write and rewrite the survey in three steps. 

\paragraph{Step 1: Draft the survey.} We first prompt an LLM to create a writing plan from the evidence papers associated with each section. The plan covers both survey-level decisions and section-level details. 
Accordingly, the LLM writes a short trial draft, which exposes overlap between sections and reveals redundancies. Guided by the trial draft, we then invoke a separate sub-agent for each section, which writes a full draft of that section.

\paragraph{Step 2: Generate figures and tables.} The draft provides the material for figures and tables, which illustrate relationships between ideas and summarize different schools of thought. For each planned figure, we invoke a sub-agent that generates multiple candidates, selects the best one, and fixes any remaining defects. 

\paragraph{Step 3: Assemble and review.} We assemble the drafted sections, figures, and tables into a complete survey, and write the introduction, conclusion, and abstract last, from the finished body. We then iteratively review and revise each part until the review no longer produces major comments.

\section{Experiments}
\subsection{Experiment Setup}
\label{sec:setup}
\begin{table*}[t]
  \centering
  \small
  \setlength{\tabcolsep}{2.4pt}
  \resizebox{\textwidth}{!}{%
    \begin{tabular}{lcccccccccccccccccccccc}
      \toprule
                           & \multicolumn{4}{c}{\textbf{Evidence}} & \multicolumn{3}{c}{\textbf{Coverage}} & \multicolumn{4}{c}{\textbf{Synthesis}} & \multicolumn{3}{c}{\textbf{New insight}} & \multicolumn{5}{c}{\textbf{Presentation}} & \multicolumn{3}{c}{\textbf{Overall}}                                                                                                                                                                                                                                                                                                                                                                                                  \\
      \cmidrule(lr){2-5} \cmidrule(lr){6-8} \cmidrule(lr){9-12} \cmidrule(lr){13-15} \cmidrule(lr){16-20} \cmidrule(lr){21-23}
      \textbf{System}      & \textit{E1}                           & \textit{E2}                           & \textit{E3}                            & \textbf{E}                               & \textit{C1}                               & \textit{C2}                          & \textbf{C}       & \textit{S1}               & \textit{S2}               & \textit{S3}               & \textbf{S}       & \textit{N1}               & \textit{N2}               & \textbf{N}       & \textit{P1}               & \textit{P2}               & \textit{P3}               & \textit{P4}               & \textbf{P}       & \textbf{Q-txt}   & \textbf{Q}       & \textbf{DAS-Eval} \\
      \midrule
      \multicolumn{23}{l}{\textit{Released surveys$^\dagger$}}                                                                                                                                                                                                                                                                                                                                                                                                                                                                                                                                                                                                                     \\
      DAS                  & \textit{63.5}                         & \textit{89.6}                         & \textit{93.3}                          & 77.5                                     & \textit{77.8}                             & \textit{59.1}                        & 68.5             & \textit{70.7}             & \textit{52.7}             & \textit{69.6}             & 64.1             & \textit{46.4}             & \textit{83.9}             & 55.8             & \textit{75.0}             & \textit{51.8}             & \textit{46.4}             & \textit{58.9}             & 58.0             & 65.6             & 64.8             & 4.45             \\
      InteractiveSurv.     & \textit{31.4}                         & \textit{17.0}                         & \textit{62.1}                          & 35.5                                     & \textit{81.8}                             & \textit{38.5}                        & 60.1             & \textit{63.7}             & \textit{52.7}             & \textit{75.9}             & 62.9             & \textit{36.6}             & \textit{91.1}             & 50.2             & \textit{73.2}             & \textit{56.2}             & \textit{25.9}             & \textit{31.2}             & 46.7             & 52.5             & 51.1             & 3.62             \\
      \midrule
      \multicolumn{23}{l}{\textit{General agent}}                                                                                                                                                                                                                                                                                                                                                                                                                                                                                                                                                                                                                                  \\
      Codex                & \textit{85.8}                         & \textit{64.4}                         & \textit{96.5}                          & 83.1                                     & \textit{\underline{91.4}}                 & \textit{\textbf{87.5}}               & \underline{89.5} & \textit{76.7}             & \textit{84.7}             & \textit{\underline{99.3}} & 85.2             & \textit{68.8}             & \textit{\underline{99.3}} & 76.4             & \textit{76.4}             & \textit{63.9}             & \textit{\underline{54.2}} & \textit{66.7}             & \underline{65.3} & 80.6             & 79.9             & \underline{4.51} \\
      \midrule
      \multicolumn{23}{l}{\textit{AutoSurvey systems}}                                                                                                                                                                                                                                                                                                                                                                                                                                                                                                                                                                                                                      \\
      TargetedRAG          & \textit{95.8}                         & \textit{\underline{89.7}}             & \textit{\underline{99.0}}              & 95.1                                     & \textit{88.0}                             & \textit{80.9}                        & 84.4             & \textit{\textbf{81.8}}    & \textit{88.2}             & \textit{\underline{99.3}} & 88.4             & \textit{66.7}             & \textit{97.2}             & 74.3             & \textit{81.2}             & \textit{\underline{73.6}} & \textit{28.5}             & \textit{\underline{70.1}} & 63.4             & 83.5             & 81.1             & \underline{4.51} \\
      AutoSurvey           & \textit{93.6}                         & \textit{84.0}                         & \textit{98.5}                          & 92.5                                     & \textit{85.6}                             & \textit{76.2}                        & 80.9             & \textit{66.2}             & \textit{97.2}             & \textit{\underline{99.3}} & 85.3             & \textit{\underline{71.5}} & \textit{\textbf{100.0}}   & \underline{78.6} & \textit{\underline{93.8}} & \textit{56.2}             & \textit{1.4}              & \textit{54.9}             & 51.6             & 81.1             & 77.8             & 4.10             \\
      SurveyForge          & \textit{93.1}                         & \textit{82.2}                         & \textit{98.4}                          & 91.7                                     & \textit{91.1}                             & \textit{84.8}                        & 87.9             & \textit{76.9}             & \textit{\textbf{98.6}}    & \textit{\underline{99.3}} & \underline{90.1} & \textit{\textbf{79.9}}    & \textit{98.6}             & \textbf{84.5}    & \textit{88.9}             & \textit{49.3}             & \textit{2.1}              & \textit{68.1}             & 52.1             & \underline{84.6} & \underline{81.3} & 4.05             \\
      SurveyAgent-HKA      & \textit{\textbf{98.7}}                & \textit{\textbf{96.6}}                & \textit{\textbf{99.1}}                 & \textbf{98.3}                            & \textit{79.1}                             & \textit{79.6}                        & 79.4             & \textit{62.5}             & \textit{86.8}             & \textit{97.2}             & 79.7             & \textit{53.5}             & \textit{97.9}             & 64.6             & \textit{79.2}             & \textit{50.0}             & \textit{4.2}              & \textit{68.1}             & 50.3             & 77.5             & 74.4             & 4.04             \\
      \midrule
      \textsc{SurveyWeave} & \textit{\underline{97.6}}             & \textit{\textbf{96.6}}                & \textit{98.9}                          & \underline{97.7}                         & \textit{\textbf{98.2}}                    & \textit{\underline{86.9}}            & \textbf{92.5}    & \textit{\underline{77.8}} & \textit{\underline{97.9}} & \textit{\textbf{100.0}}   & \textbf{90.4}    & \textit{70.8}             & \textit{\underline{99.3}} & 78.0             & \textit{\textbf{95.1}}    & \textit{\textbf{75.0}}    & \textit{\textbf{75.7}}    & \textit{\textbf{70.8}}    & \textbf{79.2}    & \textbf{87.8}    & \textbf{87.5}    & \textbf{4.58}    \\
      \bottomrule
    \end{tabular}}
  \caption{\textsc{SurveyWeave} achieves the highest \textsc{WeaveEval} and DAS-Eval scores. Italic columns are sub-metrics (Appendix~\ref{app:eval-metrics}). Q-txt excludes P3 from P. Bold: best; underline: second. $^\dagger$28 DAS-Bench topics; older models and corpora.}
  \label{tab:main}
\end{table*}

\paragraph{Data.}
We run \textsc{SurveyWeave} on Computer Science (CS) and ML-for-science topics from DAS-Bench~\cite{das} and on CS topics from Bloom-Eval~\cite{bloom}. To reduce runtime, we pre-collected 294K AI papers, mostly from four arXiv categories (cs.CL, cs.LG, cs.AI, and cs.IR) published through September 2026, and generated a summary of each paper with an LLM. Because our corpus mainly covers these categories, we exclude topics with little literature in it. This leaves 36 topics: 28 from DAS-Bench and 10 from Bloom-Eval (2 overlaps), listed in Appendix~\ref{app:topics}. Writing and evaluating 50-page surveys is very expensive, so prior benchmarks also scored and compared on small topic sets (30 in DAS-Bench and 20 in Bloom-Eval); we are comparable in scale.

\paragraph{Implementation.}
All \textsc{SurveyWeave} agents use GPT-6.1 Sol with medium reasoning effort within the Codex SDK. In paper collection, GPT-6 Luna produces relevance judgments and paper summaries. GPT-Image-2.5-Sunburst generates figures. Appendix~\ref{app:impl} lists all hyperparameters.

\paragraph{Baselines.}
We compare against three groups of systems that, wherever possible, use the same input and paper corpus as \textsc{SurveyWeave}. Every system writes one survey per topic, and we report means over topics. Appendix~\ref{app:baselines} provides details of each baseline. 

First, for the 28 DAS-Bench topics, DAS-Bench releases surveys generated by \textit{DAS} and \textit{InteractiveSurvey}~\cite{wen2025interactivesurvey}; these surveys were generated with older models and corpora. Second, we prompt the general-purpose coding agent \textit{Codex} (GPT-6.1 Sol) to search the Web and write a survey, to test whether a strong general agent already suffices.

Third, we run four AutoSurvey systems---\textit{TargetedRAG}, \textit{AutoSurvey}, \textit{SurveyForge}, and \textit{SurveyAgent-HKA}---using GPT-6.1 Sol at medium reasoning effort. \textit{TargetedRAG} follows the baseline protocol of DAS-Bench: it drafts a survey directly from the titles and abstracts of 300 papers from our paper collection, without tools or full text. \textit{SurveyForge}~\cite{yan2025surveyforge} derives outlines from its released index of human-written surveys. \textit{SurveyAgent-HKA}~\cite{surveyagenthka} runs on the abstracts in our corpus together with its own arXiv search. For every system we run, we set the target length to 15,000--22,000 words. Although we do not compare with closed-source AutoSurvey systems, their features are mostly a subset of \textit{SurveyAgent-HKA}, which is included in the comparison.

\paragraph{Metrics.}
We propose \textsc{WeaveEval}, which combines reference-free judgments with judgments grounded in multiple prior surveys. \textsc{WeaveEval} scores five dimensions on a 0--100 scale: \textit{evidence reliability} (E), whether claims are cited and supported by the cited papers; \textit{coverage} (C), whether the survey explains the field's core concepts, mechanisms, and evaluation methods; \textit{synthesis} (S), whether it relates findings, explains trade-offs, and appraises evidence critically; \textit{new insight} (N), whether it adds a perspective beyond prior surveys and proposes a specific research agenda; and \textit{presentation} (P), its organization, prose, figures, tables, and references. The overall score Q is the mean of the five dimensions. GPT-6 Luna performs claim checks and checklists, and GPT-6.1 Sol performs rubric ratings. Appendix~\ref{app:eval-metrics} defines each metric, including its weight and prompt.


We also report the DAS-Eval metrics, which score 1--5 on four dimensions: citation quality, taxonomic synthesis, hierarchical drafting, and manuscript reliability (references, figures, layout, and completeness). Following DAS-Bench, we use Qwen3.5-397B as the judge; it reads the abstracts of cited papers and judges the rendered surveys. 

\subsection{Results and Analysis}
\label{sec:results}
Table~\ref{tab:main} compares \textsc{SurveyWeave} with all baselines. It achieves the highest \textsc{WeaveEval} score (87.5), 6.2 above the next-best system (81.3). It also achieves the highest DAS-Eval score (4.58). Across the five dimensions, \textsc{SurveyWeave} scores highest on coverage (93), synthesis (90), and presentation (79), second highest on evidence reliability (98), and third on new insight (78). This reflects the benefit of precise paper collection and a well-structured outline. The lower score on new insight (78) shows room for improvement.

The released surveys score lowest: their low evidence and coverage scores (E, C) reflect weaker paper sets, and their low synthesis and new-insight scores (S, N) likely reflect the older models used to generate them. Codex, the general-purpose agent, falls short mainly on evidence reliability (83), especially on attributing statements that need a citation (64), because it does not collect papers at survey scale. The AutoSurvey baselines score within 7 points of each other overall (note that TargetedRAG uses our paper collection results). Several baselines do not generate figures and therefore score near zero on P3. For a fairer comparison, we also report Q-txt, which excludes P3; \textsc{SurveyWeave} still achieves the highest Q-txt (88), ahead of SurveyForge (85). The lead holds topic by topic and under judge re-scoring (Appendix~\ref{app:sensitivity}), and a small human evaluation is in Appendix~\ref{app:human-eval}.

\paragraph{Cost.} Generating one survey with \textsc{SurveyWeave} costs \$27.8 on average at API list prices and takes about 113 minutes, against \$0.4--\$7.7 and 12--79 minutes for the baselines. The additional cost goes mostly to reading full texts and to the agents' reviews of each other's work, and it buys the aforementioned quality gains while still taking under two hours on average. Appendix~\ref{app:cost} breaks down cost and time per system and per agent.

\subsection{Ablation Study}
\label{sec:ablation}
We ablate our agentic design. \textit{One-way pipeline} runs the tasks as a pipeline, with review and revision only within each task. In \textit{One-shot outline}, the Outline agent performs no review or revision, while the Paper and Survey agents still check in with each other. \textit{Collection-only Paper} does the same for the Paper agent. 
See Appendix~\ref{app:ablations}.

Every variant scores below the full system overall (Table~\ref{tab:ablation}). Without the Outline agent's check-ins, synthesis drops by 8.5 points. Without the Paper agent's check-ins, coverage drops by 7.3 points and synthesis by 7.4. The one-way pipeline, relying on self-checking alone, scores higher overall than the other two variants but still loses 6.8 points on evidence reliability and 6.3 on presentation.

\begin{table}[t]
  \centering
  \small
  \setlength{\tabcolsep}{3pt}
  \begin{tabular}{lcccccc}
    \toprule
    \textbf{Variant}      & \textbf{Evi}  & \textbf{Cov}  & \textbf{Syn}  & \textbf{New}  & \textbf{Pres} & \textbf{Q}    \\
    \midrule
    \textsc{SurveyWeave}  & \textbf{97.8} & \textbf{91.3} & \textbf{91.0} & \textbf{81.3} & \textbf{81.3} & \textbf{88.5} \\
    One-way pipeline      & 91.0          & 88.4          & 88.1          & 77.5          & 75.0          & 84.0          \\
    One-shot outline      & 90.9          & 86.7          & 82.5          & 80.0          & 72.5          & 82.5          \\
    Collection-only Paper & 90.1          & 84.0          & 83.6          & 77.5          & 73.8          & 81.8          \\
    \bottomrule
  \end{tabular}
  \caption{Effect of removing cross-agent check-ins.}
  \label{tab:ablation}
\end{table}



\section{Conclusion}
We presented \textsc{SurveyWeave}, an agentic architecture for automated survey generation. It decomposes automated survey generation into four tasks (paper collection, outline generation, paper categorization, and survey writing) and coordinates them with an agentic system that allows back-and-forth writing and rewriting. On topics from DAS-Bench and Bloom-Eval, \textsc{SurveyWeave} scores higher overall than every baseline, with the largest gains in coverage and presentation.

\clearpage
\section*{Limitations}
\textsc{SurveyWeave} still trails the strongest baselines in new insight, so offering perspectives that go beyond prior surveys remains open.
\textsc{SurveyWeave} costs more than the AutoSurvey baselines: about \$28 and 113 minutes per survey, because its agents read full texts and review each other's work. We leave improving efficiency to future work. 

Because each agent owns an entity, a natural extension is to keep a survey up to date as new papers appear, revising only the affected sections rather than regenerating it. Additionally, \textsc{SurveyWeave} requires input about which surveys to write. Another extension is to mine the whole paper corpus to identify areas where surveys are worthwhile.

Our main evaluation uses LLM judges, including GPT-6.1 Sol, the same model that drives \textsc{SurveyWeave} and the baselines. We partly address this by also reporting DAS-Eval, whose judge is a different model.

Finally, our comparison with baselines focuses on computer science and machine learning for science, and all topics and generated surveys are in English. Generalization to other fields will be an important future direction to expand the impact of \textsc{SurveyWeave}. 

\section*{Ethics Statement}
Generated surveys can contain unsupported or misattributed claims, and fluent text may lead readers to trust them more than they should. \textsc{SurveyWeave}'s surveys are meant as a starting point for researchers, not a substitute for reading the cited work; our evidence-reliability metrics measure how often claims are unsupported. All language models in this work are accessed through hosted APIs, with the DAS-Eval judge served through OpenRouter, and we trained no language models. The benchmarks we use are released under open licenses (DAS-Bench under CC BY 4.0 and Bloom-Eval under MIT), and we use them and the baseline systems' released code for research only. We used AI assistants to help write code and edit this paper; the authors reviewed all changes and take responsibility for the content.


\bibliography{custom}

\begin{thebibliography}{24}
\providecommand{\natexlab}[1]{#1}

\bibitem[{Bao et~al.(2026)Bao, Nayeem, Zhao, Rafiei, and
  Zhang}]{surveyagenthka}
Tong Bao, Mir~Tafseer Nayeem, Yi~Zhao, Davood Rafiei, and Chengzhi Zhang. 2026.
\newblock \href {https://doi.org/10.1016/j.knosys.2026.117011}
  {{SurveyAgent-HKA}: A multi-agent framework for scientific survey generation
  with {LLMs} and human knowledge augmentation}.
\newblock \emph{Knowledge-Based Systems}, 352:117011.

\bibitem[{Chao et~al.(2025)Chao, Lin, Wang, Zhang, Zhou, Wang, Wang, Zhou, Liu,
  and Sun}]{mapreducev3}
Yu~Chao, Siyu Lin, Xiaorong Wang, Zhu Zhang, Zihan Zhou, Haoyu Wang, Shuo Wang,
  Jie Zhou, Zhiyuan Liu, and Maosong Sun. 2025.
\newblock \href {https://doi.org/10.18653/v1/2025.emnlp-demos.51}
  {{LLM}$\times${MapReduce}-{V3}: Enabling interactive in-depth survey
  generation through a {MCP}-driven hierarchically modular agent system}.
\newblock In \emph{Proceedings of the 2025 Conference on Empirical Methods in
  Natural Language Processing: System Demonstrations}, pages 688--695, Suzhou,
  China. Association for Computational Linguistics.

\bibitem[{Chen et~al.(2025)Chen, Yang, Shen, Liu, Belloum, Grosso, and
  Papagianni}]{surveygeni}
Jing Chen, Zhiheng Yang, Yixian Shen, Jie Liu, Adam Belloum, Paola Grosso, and
  Chrysa Papagianni. 2025.
\newblock \href {https://doi.org/10.18653/v1/2025.ijcnlp-long.193}
  {{SurveyGen-I}: Consistent scientific survey generation with evolving plans
  and memory-guided writing}.
\newblock In \emph{Proceedings of the 14th International Joint Conference on
  Natural Language Processing and the 4th Conference of the Asia-Pacific
  Chapter of the Association for Computational Linguistics}, pages 3687--3714,
  Mumbai, India. The Asian Federation of Natural Language Processing and The
  Association for Computational Linguistics.

\bibitem[{Cormack et~al.(2009)Cormack, Clarke, and
  B{\"u}ttcher}]{cormack2009rrf}
Gordon~V. Cormack, Charles L.~A. Clarke, and Stefan B{\"u}ttcher. 2009.
\newblock \href {https://doi.org/10.1145/1571941.1572114} {Reciprocal rank
  fusion outperforms condorcet and individual rank learning methods}.
\newblock In \emph{Proceedings of the 32nd International ACM SIGIR Conference
  on Research and Development in Information Retrieval}, pages 758--759.

\bibitem[{{Google}(2024)}]{google2024deepresearch}
{Google}. 2024.
\newblock {Gemini}: Try {Deep Research} and {Gemini} 2.0 {Flash Experimental}.
\newblock
  \url{https://blog.google/products/gemini/google-gemini-deep-research/}.

\bibitem[{Huang et~al.(2025)Huang, Chen, Zhang, Li, Zhou, Fang, Yang, Li,
  Shang, Xu, Hao, Shao, and Wang}]{huang2025deepresearchagents}
Yuxuan Huang, Yihang Chen, Haozheng Zhang, Kang Li, Huichi Zhou, Meng Fang,
  Linyi Yang, Xiaoguang Li, Lifeng Shang, Songcen Xu, Jianye Hao, Kun Shao, and
  Jun Wang. 2025.
\newblock Deep research agents: A systematic examination and roadmap.
\newblock \emph{arXiv preprint arXiv:2506.18096}.

\bibitem[{Jiang et~al.(2024)Jiang, Shao, Ma, Semnani, and
  Lam}]{jiang2024costorm}
Yucheng Jiang, Yijia Shao, Dekun Ma, Sina Semnani, and Monica Lam. 2024.
\newblock \href {https://doi.org/10.18653/v1/2024.emnlp-main.554} {Into the
  unknown unknowns: Engaged human learning through participation in language
  model agent conversations}.
\newblock In \emph{Proceedings of the 2024 Conference on Empirical Methods in
  Natural Language Processing}, pages 9917--9955, Miami, Florida, USA.
  Association for Computational Linguistics.

\bibitem[{Liang et~al.(2025)Liang, Yang, Wang, Tang, Zheng, Song, Lin, Yang,
  Niu, Wang, Tang, Xiong, Mao, and Li}]{liang2025surveyx}
Xun Liang, Jiawei Yang, Yezhaohui Wang, Chen Tang, Zifan Zheng, Shichao Song,
  Zehao Lin, Yebin Yang, Simin Niu, Hanyu Wang, Bo~Tang, Feiyu Xiong, Keming
  Mao, and Zhiyu Li. 2025.
\newblock {SurveyX}: Academic survey automation via large language models.
\newblock \emph{arXiv preprint arXiv:2502.14776}.

\bibitem[{{OpenAI}(2025)}]{openai2025deepresearch}
{OpenAI}. 2025.
\newblock Introducing deep research.
\newblock \url{https://openai.com/index/introducing-deep-research/}.

\bibitem[{{Perplexity}(2025)}]{perplexity2025deepresearch}
{Perplexity}. 2025.
\newblock Introducing {Perplexity Deep Research}.
\newblock
  \url{https://www.perplexity.ai/hub/blog/introducing-perplexity-deep-research}.

\bibitem[{Shao et~al.(2024)Shao, Jiang, Kanell, Xu, Khattab, and
  Lam}]{shao2024storm}
Yijia Shao, Yucheng Jiang, Theodore Kanell, Peter Xu, Omar Khattab, and Monica
  Lam. 2024.
\newblock \href {https://doi.org/10.18653/v1/2024.naacl-long.347} {Assisting in
  writing {W}ikipedia-like articles from scratch with large language models}.
\newblock In \emph{Proceedings of the 2024 Conference of the North American
  Chapter of the Association for Computational Linguistics: Human Language
  Technologies (Volume 1: Long Papers)}, pages 6252--6278, Mexico City, Mexico.
  Association for Computational Linguistics.

\bibitem[{Shi et~al.(2025)Shi, Kou, Li, Tang, Xie, Yu, Wang, and
  Zhou}]{scisage}
Xiaofeng Shi, Qian Kou, Yuduo Li, Ning Tang, Jinxin Xie, Longbin Yu, Songjing
  Wang, and Hua Zhou. 2025.
\newblock {SciSage}: A multi-agent framework for high-quality scientific survey
  generation.
\newblock \emph{arXiv preprint arXiv:2506.12689}.

\bibitem[{S{\o}rensen(1948)}]{sorensen1948method}
Thorvald S{\o}rensen. 1948.
\newblock A method of establishing groups of equal amplitude in plant sociology
  based on similarity of species content and its application to analyses of the
  vegetation on {Danish} commons.
\newblock \emph{Biologiske Skrifter}, 5(4):1--34.

\bibitem[{Wang et~al.(2025)Wang, Fu, Zhang, Wang, Ren, Wang, Li, He, An, Liu,
  and Sun}]{wang2025mapreducev2}
Haoyu Wang, Yujia Fu, Zhu Zhang, Shuo Wang, Zirui Ren, Xiaorong Wang, Zhili Li,
  Chaoqun He, Bo~An, Zhiyuan Liu, and Maosong Sun. 2025.
\newblock {LLM}$\times${MapReduce}-{V2}: Entropy-driven convolutional test-time
  scaling for generating long-form articles from extremely long resources.
\newblock \emph{arXiv preprint arXiv:2504.05732}.

\bibitem[{Wang et~al.(2024)Wang, Guo, Yao, Zhang, Zhang, Wu, Zhang, Dai, Zhang,
  Wen, Ye, Zhang, and Zhang}]{wang2024autosurvey}
Yidong Wang, Qi~Guo, Wenjin Yao, Hongbo Zhang, Xin Zhang, Zhen Wu, Meishan
  Zhang, Xinyu Dai, Min Zhang, Qingsong Wen, Wei Ye, Shikun Zhang, and Yue
  Zhang. 2024.
\newblock {AutoSurvey}: Large language models can automatically write surveys.
\newblock In \emph{Advances in Neural Information Processing Systems},
  volume~37.

\bibitem[{Wen et~al.(2025)Wen, Cao, Wang, Guo, Yang, and
  Liu}]{wen2025interactivesurvey}
Zhiyuan Wen, Jiannong Cao, Zian Wang, Beichen Guo, Ruosong Yang, and Shuaiqi
  Liu. 2025.
\newblock {InteractiveSurvey}: An {LLM}-based personalized and interactive
  survey paper generation system.
\newblock \emph{arXiv preprint arXiv:2504.08762}.

\bibitem[{Xu and Peng(2025)}]{xu2025deepresearchsurvey}
Renjun Xu and Jingwen Peng. 2025.
\newblock A comprehensive survey of deep research: Systems, methodologies, and
  applications.
\newblock \emph{arXiv preprint arXiv:2506.12594}.

\bibitem[{Xu et~al.(2026{\natexlab{a}})Xu, Zhao, Wei, Kou, Han, and
  Wei}]{fiksurvey}
Tianyi Xu, Zhe Zhao, Tianshuo Wei, Yiqun Kou, Liuliu Han, and Ye~Wei.
  2026{\natexlab{a}}.
\newblock Feedback is the key for automated survey generation.
\newblock In \emph{Findings of the Association for Computational Linguistics:
  ACL 2026}, pages 38174--38195, San Diego, California, United States.
  Association for Computational Linguistics.

\bibitem[{Xu et~al.(2026{\natexlab{b}})Xu, Xue, Hu, Wang, Liu, and Zhang}]{das}
Zhikai Xu, Zhucun Xue, Teng Hu, Yabiao Wang, Yong Liu, and Jiangning Zhang.
  2026{\natexlab{b}}.
\newblock Deep academic survey: Stateful agentic closed-loop paradigm for
  academic survey automation.
\newblock \emph{arXiv preprint arXiv:2608.18034}.

\bibitem[{Yan et~al.(2025)Yan, Feng, Yuan, Xia, Wang, Bai, and
  Zhang}]{yan2025surveyforge}
Xiangchao Yan, Shiyang Feng, Jiakang Yuan, Renqiu Xia, Bin Wang, Lei Bai, and
  Bo~Zhang. 2025.
\newblock \href {https://doi.org/10.18653/v1/2025.acl-long.609} {{SurveyForge}:
  On the outline heuristics, memory-driven generation, and multi-dimensional
  evaluation for automated survey writing}.
\newblock In \emph{Proceedings of the 63rd Annual Meeting of the Association
  for Computational Linguistics (Volume 1: Long Papers)}, pages 12444--12465,
  Vienna, Austria. Association for Computational Linguistics.

\bibitem[{Yang et~al.(2026)Yang, Ma, Li, Wang, Huang, Hu, Wang, Zhang, Wu, Yu,
  and Chen}]{deepsurvey}
Ziyue Yang, Da~Ma, Hanqi Li, Zijian Wang, Tiancheng Huang, Zijian Hu, Chenrun
  Wang, Yunzhe Zhang, Xiaobao Wu, Kai Yu, and Lu~Chen. 2026.
\newblock {DeepSurvey}: Agent-oriented automated survey generation with
  analytical depth and citation reliability.
\newblock \emph{arXiv preprint arXiv:2605.29522}.

\bibitem[{Zhang et~al.(2026)Zhang, Zhao, Wen, Wei, Zhang, Yang, and
  Wei}]{bloom}
Fei Zhang, Zhe Zhao, HaiBin Wen, Tianshuo Wei, Zaixi Zhang, Chao Yang, and
  Ye~Wei. 2026.
\newblock \href {https://doi.org/10.18653/v1/2026.acl-long.1315} {{Bloom-Eval}:
  A hierarchical evaluation benchmark for automatic survey generation based on
  {Bloom}'s taxonomy}.
\newblock In \emph{Proceedings of the 64th Annual Meeting of the Association
  for Computational Linguistics (Volume 1: Long Papers)}, pages 28512--28544.

\bibitem[{Zhang et~al.(2025)Zhang, Cui, Wang, Tian, Guo, Wang, Wu, Song, and
  Zhang}]{itersurvey}
Hongbo Zhang, Han Cui, Yidong Wang, Yijian Tian, Qi~Guo, Cunxiang Wang, Jian
  Wu, Chiyu Song, and Yue Zhang. 2025.
\newblock Deep literature survey automation with an iterative workflow.
\newblock \emph{arXiv preprint arXiv:2510.21900}.

\bibitem[{Zheng et~al.(2025)Zheng, Deng, Tsang, Wang, Bai, Wang, and
  Song}]{zheng2025automation}
Tianshi Zheng, Zheye Deng, Hong~Ting Tsang, Weiqi Wang, Jiaxin Bai, Zihao Wang,
  and Yangqiu Song. 2025.
\newblock From automation to autonomy: A survey on large language models in
  scientific discovery.
\newblock In \emph{Proceedings of the 2025 Conference on Empirical Methods in
  Natural Language Processing}, pages 17744--17761.

\end{thebibliography}

\clearpage
\beginappendix

\renewcommand{\topfraction}{0.9}
\renewcommand{\bottomfraction}{0.8}
\renewcommand{\textfraction}{0.07}
\renewcommand{\floatpagefraction}{0.85}
\renewcommand{\dbltopfraction}{0.9}
\renewcommand{\dblfloatpagefraction}{0.85}
\setcounter{topnumber}{4}
\setcounter{totalnumber}{6}
\setcounter{dbltopnumber}{3}

\section{Algorithm}
\label{app:algorithm}
\begin{algorithm}[H]
  \caption{\textsc{SurveyWeave}.}
  \label{alg:surveyweave}
  \begin{algorithmic}[1]
    \Require topic $T$, paper database
    \Ensure survey $S$
    \State \textbf{Paper:} collect $P$ with $\mathcal{C}(p)$ and prior surveys $P_{\mathrm{surv}}$; map their coverage; send to Outline
    \State \textbf{Outline:} draft $O$; \textbf{Paper} reviews it for coverage and evidence per section; finalize $O$
    \State \textbf{Paper:} categorize papers into $E(v)$ for every $v \in V$
    \For{each section $v$ still lacking evidence}
    \State \textbf{Outline:} cut, merge, narrow, or keep $v$ short; \textbf{Paper} updates $E$
    \EndFor
    \State \textbf{Survey:} write a plan from $O$, $E$ and $P_{\mathrm{surv}}$; \textbf{Paper} and \textbf{Outline} review it
    \State \textbf{Survey:} draft $S$; literature questions go to Paper, structure problems to Outline
    \State \quad if $O$ changes, Paper updates $E$ and Survey rewrites the changed sections
    \Repeat
    \State \textbf{Paper}, \textbf{Outline} and \textbf{Survey} review $S$ and record issues
    \State \textbf{Survey:} revise $S$; each reviewer verifies the fixes to its own issues
    \Until{no major issues remain, or three rounds}
  \end{algorithmic}
\end{algorithm}

\section{Implementation Details}
\label{app:impl}
\label{appendix:implementation}
This section gives the settings needed to reproduce \textsc{SurveyWeave} as described in \S\ref{sec:surveyweave}: models, the paper database, the agent runtime and workbench, the sub-agents, and the details of each task. Table~\ref{tab:hyperparams} lists the hyperparameters. All values were set heuristically; values marked $^\dagger$ are guidance in the agents' instructions rather than limits enforced in code.

\paragraph{Models.} The three agents and all their sub-agents use GPT-6.1 Sol with medium reasoning effort. In paper collection, GPT-6 Luna (medium effort) writes the search queries, judges each paper's relevance, and writes its description and concept tags. GPT-6 Luna (medium effort) also writes the structured summary of any corpus paper that lacks one. Figures are generated with GPT-Image-2.5-Sunburst. Semantic search over the database uses \texttt{text-embedding-3-small} (512 dimensions); section candidate ranking and concept clustering use \texttt{gemini-embedding-001} (3,072 dimensions).

\paragraph{Paper database.} To reduce runtime, we pre-collected 294K AI papers (294,043 records), mostly from four arXiv categories (cs.CL, cs.LG, cs.AI and cs.IR) published from 2023 through September 2026, and froze the database before the runs. Like arXiv as a whole, the corpus is predominantly English; all topics, prompts, and generated surveys are in English. Each record holds the title, abstract, authors, year, venue and link, an abstract embedding, and a structured summary generated by an LLM. A summary has five fields: basics, core idea, technical details, experiments, and figures; summaries generated during a run read at most 50,000 characters of the paper's text. A fixed snapshot of a citation graph supplies references and overall citation counts. Agents read a paper through its summary; the Paper agent and its sub-agents can also read its full text.

\paragraph{Agent runtime.} The agents run as concurrent threads in the Codex SDK. A host thread starts the Paper, Outline and Survey agents, waits while they work, and makes no decision about the survey. It sends a notice if all three agents are idle while the survey is unfinished. Agents communicate through two tools: a \textit{task} tool hands a teammate work or a question and starts its turn, and a \textit{message} tool delivers an update that needs no action. A message names the workbench files it concerns rather than copying their content. On the 36 topics, a run made on average 75 task calls and 123 message calls.

\paragraph{Workbench.} The workbench is a directory shared by all agents and sub-agents, plus a state file that points to the current corpus. It holds the topic $T$ (topic name, description and a main-text range of 15,000--22,000 words); the corpus ${\bf P}$ as a versioned JSON file, with the collection's concept tags and citation counts; the Paper agent's coverage map of ${\bf P}$ and its profiles of the prior surveys ${\bf P}_{\mathrm{surv}}$; the outline ${\bf O}$ with each section's title, scope and a rationale (the Outline agent's files); the evidence sets ${\bf E}$ as one JSON file listing, for every section, its papers with a one-line role each and the paths of notes about them; the writing plan; and the \LaTeX{} source, figures and compiled builds of $S$. Each artifact has one owner agent that may change it; the others send change requests to the owner. Reviews go into one append-only ledger. Each entry records who raised it, where the problem is, what is wrong and why it matters to a reader, whether it is major or minor, and the recommended fix. The Survey agent then writes how it resolved the entry, and the reviewer marks it resolved or reopens it.

\paragraph{Sub-agents.} An agent spawns a sub-agent as a fresh thread that receives only its task instructions and the paths it needs, and writes its reasoning to its own directory. At most 12 sub-agent threads are open at once across the team. On the 36 topics a run used about 34 sub-agents on average (29--43). The Paper agent uses them to match papers to groups of sections (3.8 per run), to answer literature questions and check figure and table facts from full text (7.7), to check the survey's claims during review (1.8), and to profile prior surveys (1.5). The Outline agent uses research sub-agents that write the two outline-evidence reports, two sub-agents that compare outline alternatives, and an independent outline reviewer (about 4 in total). The Survey agent uses one section author per top-level body section and for the introduction, conclusion, title and abstract (8.7), one figure maker per figure plus repairs (4.8), and one visual reviewer.

\paragraph{Paper collection.} GPT-6 Luna writes 3--4 semantic descriptions and 4--8 keyword queries for the topic; a keyword query is a conjunction of 1--3 literal phrases matched in titles and abstracts. Each semantic description retrieves its 1,500 nearest papers by embedding, and the result lists of all queries are merged by reciprocal-rank fusion~\cite{cormack2009rrf} (constant 60). The distilled ranker averages the probabilities of two logistic regressions, one on TF-IDF features of the title and abstract and one on abstract embeddings. It is trained on LLM labels for the top 200 and 200 random query results, then refit after labeling the 100 highest-scoring papers outside the query results. Unjudged query results are judged first, then the rest of the database in ranker order. For each paper, the judge returns core, not core or uncertain, the authors' aim, the contribution, a reason, sentence IDs from the abstract supporting a core label, and 1--3 concept tags, each with a short definition. A core label without valid supporting sentences becomes uncertain. The prompt asks the judge to respect restrictive modifiers in the topic description and to count a paper as core only if the topic is what it advances, not a tool it uses for another goal. Judging stops when two consecutive batches have fewer than 4\% core papers among completed judgments (the first four batches of query results never count toward this), or at 30,000 judgments. A random sample of up to 100 unjudged papers is then judged to recover core papers below the cutoff. For background papers, the collector follows the references of the core papers in the citation graph, recovers up to 250 of the most-cited references missing from the database, and selects up to $B = \lceil 16\sqrt{|{\bf P}_{\mathrm{core}}|}\,\rceil$ papers: references cited by at least two core papers first, ranked by the number of citing core papers and then overall citations, then references cited by one core paper, preferring core papers not yet represented. Each background paper also receives a description and tags, and is promoted to core if the judge finds it explicitly addresses the topic. On the 36 topics, collection judged 5,392 papers on average (3,600--8,600) and kept 1,187 core papers (324--3,015) and 395 background papers (196--617). The Paper agent then identifies prior surveys on the topic and neighboring topics, which form ${\bf P}_{\mathrm{surv}}$.

\paragraph{Outline generation.} The concept tags of the core papers, each represented by its definition, are embedded and grouped by agglomerative clustering with complete linkage and cosine distance into $K=\min(200, \mathrm{number\ of\ distinct\ tags})$ clusters. Tags of background papers are kept as a separate list, so that they do not count toward the size of a cluster. The Outline agent reads the clusters, the Paper agent's coverage map and two reports on the field's central concepts, key mechanisms and contested boundaries, one drawn from the Web and one from the prior surveys in ${\bf P}_{\mathrm{surv}}$ (\S\ref{sec:impl-outline}, Step 2). It sketches three candidate outlines, each organizing the field along a different dimension, and has a comparison sub-agent assess what each makes clear and what it obscures. It writes one outline, explores alternatives for its structure and depth with a second comparison sub-agent, and then has the draft reviewed by an independent reviewer and by the Paper agent before finalizing it with a rationale.

\paragraph{Paper categorization.} For each section and subsection, the Paper agent retrieves candidate papers that match the section's title, scope and concept tags, and ranks them by quality: by citation count for earlier papers, and, for recent papers whose citation counts are not yet informative, by the quality score that the LLM generates together with the paper summary. The agent sees the top 30 candidates with their titles, years, abstracts and quality signals, and selects 3--10 papers per section. A section is thin when fewer than about three of its papers address its scope directly, or when its scope promises a comparison its papers cannot support. For a thin section the agent pages further down the ranking, to rank 90 at most. A section still thin after that goes to the Outline agent, which cuts, merges, narrows, or keeps it short; the Paper agent then rematches only the sections whose scope changed. Finally the agent places each of the 30 core papers most cited within the core set in the section a reader would expect, and generates summaries for every selected paper.

\paragraph{Survey writing.} The Survey agent writes a plan of reader questions, the comparisons that answer them, and the visuals, with at least two figures and two tables; the Paper and Outline agents review it. From the plan, the Survey agent writes a short trial draft that exposes overlap and redundancy between sections. Guided by the trial draft, body sections are then drafted one top-level section at a time, in order, each by a fresh section author that receives the plan, the section's papers and notes, and the finished text of all earlier sections. Figures and tables are then made from the draft, with the Paper agent checking the facts each one depends on against the papers' full text. A figure maker studies a pack of reference figures redrawn from published surveys, writes a brief and three prompts, and waits for the Survey agent to approve the concept. It then generates three candidates with GPT-Image-2.5-Sunburst, each conditioned on up to four reference images, selects one for accuracy, clarity and visual quality, and repairs remaining defects by regenerating from the selected image. After the body and visuals are assembled, section authors write the introduction, conclusion, title and abstract from the finished body. The Survey agent reviews the draft itself, with a visual reviewer checking every figure and table in the rendered pages. On the 36 topics a survey had on average 3.2 figures, 3.8 tables and 157 cited papers, and a run made 19.5 image-generation calls.

\paragraph{Review rounds.} The Paper agent reviews the compiled survey for claims (checked against the cited papers' full text, starting with the results that comparisons rest on) and for coverage of the main approaches in each section's ranking. The Outline agent reviews whether each section does the job the outline gave it and whether sections overlap. A major issue is one that would mislead a reader or keep a section from doing its job; wording and style are minor. The Survey agent resolves every entry, sending evidence gaps to the Paper agent and structure problems to the Outline agent; if the outline changes, the Paper agent rematches the affected sections and the Survey agent rewrites them. Each reviewer then verifies the fixes to its own entries. The loop repeats while major entries remain open, for up to three rounds.

\paragraph{Building the PDF.} Agents compile through a host tool that inlines all included source files, copies the included images into a new build directory, and rejects any citation key that is not a corpus paper. It generates the bibliography with BibTeX (\texttt{unsrt}) from corpus metadata; agents may correct only an entry's authors, title, venue and year. It compiles with \texttt{latexmk} and \texttt{pdflatex} without shell escape, halting on the first error, and renders every page as a preview image. Each build also reports measured signals for the agents to inspect, such as paragraphs over 220 words, sentences over 50 words, papers cited more than 10 times, and fewer than two figures or tables.

\begin{table}[htbp]
  \centering
  \small
  \begin{tabular}{lr}
    \toprule
    \textbf{Hyperparameter}                      & \textbf{Value}    \\
    \midrule
    \multicolumn{2}{l}{\textit{Paper collection}}                    \\
    Semantic / keyword queries                   & 3--4 / 4--8       \\
    Semantic hits per query                      & 1,500             \\
    Classifier round 1 (top / random)            & 200 / 200         \\
    Classifier round 2 (outside queries)         & 100               \\
    Background budget coefficient                & 16                \\
    Concept tags per paper                       & 1--3              \\
    Judgment batch size                          & 100               \\
    Stop: relevant rate in 2 batches             & $<$4\%            \\
    Protected initial query batches              & 4                 \\
    Maximum judgments                            & 30,000            \\
    \midrule
    \multicolumn{2}{l}{\textit{Outline generation}}                  \\
    Maximum tag groups                           & 200               \\
    Clustering linkage / distance                & complete / cosine \\
    Candidate outlines$^\dagger$                 & 3                 \\
    \midrule
    \multicolumn{2}{l}{\textit{Paper categorization}}                \\
    Candidates per node $k$                      & 30                \\
    Candidates checked for a thin node$^\dagger$ & 90                \\
    Target evidence set size $|E(v)|$$^\dagger$  & 3--10             \\
      \midrule
    \multicolumn{2}{l}{\textit{Survey writing}}                      \\
    Minimum figures / tables$^\dagger$           & 2 / 2             \\
    Figure candidates                            & 3                 \\
    Reference images per image call              & $\le$4            \\
    Revision rounds after first review$^\dagger$ & $\le$3            \\
      \midrule
    \multicolumn{2}{l}{\textit{Runtime}}                             \\
    Open sub-agent threads                       & $\le$12           \\
    \bottomrule
  \end{tabular}
  \caption{Pipeline hyperparameters.}
  \label{tab:hyperparams}
\end{table}

\section{Baseline Configurations}
\label{app:baselines}
This section describes how we ran or obtained each baseline in Table~\ref{tab:main}: the code, input, papers, model, length settings, output format, and every change we made to released code. Table~\ref{tab:baseline-config} summarizes the configurations, and Figures~\ref{fig:prompt-targetedrag} and~\ref{fig:prompt-codex} reproduce the two task prompts that we wrote ourselves.

\paragraph{Common setup.} Every system we run receives the same input as \textsc{SurveyWeave}: the topic name and its one-sentence description (Appendix~\ref{app:topics}). The four AutoSurvey baselines read papers from the same corpus of 294K papers as \textsc{SurveyWeave} (\S\ref{sec:setup}). Codex searches the open Web. TargetedRAG, AutoSurvey, SurveyForge and SurveyAgent-HKA call GPT-6.1 Sol with medium reasoning effort. Each system writes one survey per topic. We convert the text output of TargetedRAG, AutoSurvey, SurveyForge and SurveyAgent-HKA to LaTeX and PDF with a deterministic converter that maps citations to bibliography entries and does not rewrite prose; Codex compiles its own PDF. \textsc{WeaveEval} and DAS-Eval score the PDFs.

\paragraph{Length calibration.} Where a system takes the length as an instruction (TargetedRAG and Codex), we ask for a main text of 15,000--22,000 words. The released pipelines fix length through the number of sections and a per-subsection word target, so we set those parameters so that their surveys are of comparable length: AutoSurvey writes 7 sections of 3--4 subsections at about 500 words each, SurveyForge writes 7 body sections with $\max(500, \mathrm{round}(18{,}500/S))$ words per subsection for $S$ subsections, and SurveyAgent-HKA writes at most 50 subsections of 300--500 words each. We neither truncate nor reject outputs by length; Appendix~\ref{app:sensitivity} reports the resulting lengths.

\begin{table*}[tp]
  \centering
  \footnotesize
  \setlength{\tabcolsep}{3pt}
  \begin{tabularx}{\textwidth}{@{}l >{\raggedright\arraybackslash}p{2.6cm} >{\raggedright\arraybackslash}X c >{\raggedright\arraybackslash}p{2.4cm} >{\raggedright\arraybackslash}X@{}}
    \toprule
    \textbf{System}      & \textbf{Code}                     & \textbf{Papers available}                                                          & \textbf{Full text} & \textbf{Figures / tables}          & \textbf{Length setting}                                            \\
    \midrule
    \textsc{SurveyWeave} & Ours                              & Our corpus, collected per topic, and the Web                                       & Yes                & Generated figures; tables          & 15,000--22,000 words (instruction)                                 \\
    TargetedRAG          & Ours, following DAS-Bench         & 300 papers from \textsc{SurveyWeave}'s paper collection                            & No                 & TikZ figures; tables               & 15,000--22,000 words (prompt)                                      \\
    AutoSurvey           & Released, commit \texttt{5e8f389} & Titles and abstracts of our corpus                                                 & No                 & Text only                          & 7 sections, 3--4 subsections each, about 500 words per subsection  \\
    SurveyForge          & Released, commit \texttt{9114a0b} & Titles and abstracts of our corpus; released index of 12,756 human-written surveys & No                 & Text only                          & 7 body sections; per subsection, $18{,}500/S$ words (at least 500) \\
    SurveyAgent-HKA      & Released, commit \texttt{4f544b4} & Abstracts of our corpus; live arXiv search                                         & No                 & Text; no figure step               & at most 50 subsections, 300--500 words each                        \\
    Codex                & Codex agent, our task prompt      & Open Web                                                                           & Yes                & Its own tools; no image generation & 15,000--22,000 words (prompt)                                      \\
    DAS                  & Released surveys                  & As released                                                                        & --                 & As released                        & As released                                                        \\
    InteractiveSurvey    & Released surveys                  & As released                                                                        & --                 & As released                        & As released                                                        \\
    \bottomrule
  \end{tabularx}
  \caption{Baseline configurations. All systems we run use GPT-6.1 Sol with medium reasoning effort. The released DAS and InteractiveSurvey surveys were generated by their authors with older models and corpora. $S$: number of subsections.}
  \label{tab:baseline-config}
\end{table*}

\paragraph{TargetedRAG.} TargetedRAG follows the direct-drafting baseline protocol of DAS-Bench~\cite{das}, which describes drafting a survey from a fixed pool of paper metadata but does not release its prompt; we therefore wrote the implementation and prompt ourselves. For each topic, we build a packet of 300 papers from \textsc{SurveyWeave}'s paper collection for that topic: the 100 papers most cited by the topic's core papers, and the 200 remaining papers most similar to the topic description (text-embedding-3-small, 512 dimensions), deduplicated by paper identity. Each record is numbered and gives only the title, year, authors and abstract; records are listed in year and title order. The model receives the prompt in Figure~\ref{fig:prompt-targetedrag} in a single call, with no tools and no full text. The model writes LaTeX body markup and may draw figures with TikZ or pgfplots; our packager adds the preamble and bibliography and maps record numbers to corpus entries.

\paragraph{AutoSurvey.} AutoSurvey~\cite{wang2024autosurvey} retrieves papers, drafts outlines from chunks of retrieved papers and merges them, then writes subsections in parallel and refines citations and coherence. We ran its released code (commit \texttt{5e8f389}). The released code drafts from abstracts and does not call its full-text loader, so our run also uses abstracts. We replaced its 530K-paper arXiv database and FAISS index with titles and abstracts from our corpus, retrieved by a hybrid of dense similarity (text-embedding-3-small, 512 dimensions) and keyword matching, fused with reciprocal rank $1/(20+\mathrm{rank})$. We kept the released retrieval budgets: 1,500 papers for outline generation in 30,000-token chunks, 50 papers per section when expanding subsections, and 60 papers per subsection for drafting and citation checking. Its citation reflection and its two waves of local-coherence refinement are unchanged. Our changes to the released code are: (1) length calibration in three prompts, where the merge prompt states the section count (7), the subsection prompt asks for 3--4 subsections instead of ``several'', and the drafting prompt asks for ``about'' 500 words instead of ``more than'' a word count, with the driver's per-subsection length fixed at 500; (2) two fixes to list variables that the released writer shared across subsections, so each subsection keeps its own papers and references; (3) worker failures and empty or malformed subsections now stop the run instead of leaving silent gaps; and (4) unused GPU imports removed. AutoSurvey links citations by matching cited titles to the nearest corpus title; our converter restores bracketed math and notation that this linker had mistaken for paper titles. AutoSurvey produces Markdown text without figures or tables, which we convert to LaTeX.

\paragraph{SurveyForge.} SurveyForge~\cite{yan2025surveyforge} builds outlines with guidance from a database of human-written survey outlines and retrieves papers per subsection with a memory of earlier retrievals. We ran its released code (commit \texttt{9114a0b}) with its prompt module unchanged. On the survey side, we kept its released index of human-written surveys (12,756 records, embedded with gte-large-en-v1.5) and its outline procedure: five surveys sampled from the top 20 as priors for each rough outline, and the top 10 and then the 5 most recent as priors for merging. On the paper side, we replaced its paper database with titles and abstracts from our corpus (exact inner-product search over 512-dimensional embeddings) and its citation counts with Semantic Scholar counts. We kept the 1,500-paper memory, the top 100 papers per sub-query, the reranking by citations within two-year windows down to 60 papers per subsection, and 50 sampled papers per section sub-outline. We extended the end of its hard-coded citation windows from September 26, 2024 to our corpus cutoff; its 2012 start date has no effect, since our corpus begins in 2023. Our other changes are: the same two list-variable fixes as in AutoSurvey; worker failures stop the run; when its merge step returns no title, the topic name is used as the title; bracketed math and notation are kept rather than linked to papers; and sampling uses a fixed seed (20261009). SurveyForge produces Markdown text without figures or tables.

\paragraph{SurveyAgent-HKA.} SurveyAgent-HKA~\cite{surveyagenthka} uses human-written surveys to build outlines and peer reviews to guide its reviewer, writes subsections from hierarchical knowledge cards, and revises the draft for up to three rounds. We ran its released code (commit \texttt{4f544b4}) and passed our model, retrieval and citation-data components into its pipeline instead of rewriting it. Its Semantic Scholar search is replaced by search over the abstracts in our corpus; its live arXiv search is kept, with hits limited to the corpus cutoff. Citation counts, references and DOIs come from a local Semantic Scholar release. Its human-outline input is the SurveyGen corpus of human-written surveys, and it keeps its upstream embedder (BAAI/bge-large-en-v1.5). We used its settings of 150 initial papers, 60 results per section query, 20 co-cited papers, 50 papers per section, 6 literature clusters, 10 human outlines and 20 peer-review references. Our changes are: (1) length calibration in three prompts, asking for at most 50 subsections in the first and refined outlines (regenerating once if exceeded, never truncating) and for 300--500 words per subsection instead of at least 300; (2) the reviser returns only changed or added subsections, which are merged into the draft, and runs as four parallel passes over disjoint groups of sections, because full-draft revisions exceeded the provider's response limit; a revision that keeps less than 85\% of the draft's words is rejected, and after three failed attempts the round keeps the unrevised draft, which the upstream scorer then accepts or rejects; (3) front matter, the survey's own methodology section and its limitations section are exempt from the check that every subsection cites a paper; and (4) its Markdown output is converted to LaTeX with the converter used for SurveyForge, mapping Unicode math symbols to LaTeX. Its upstream scorer and stop-if-no-improvement rule are unchanged.

\paragraph{Codex.} Codex is a general-purpose coding agent. We run it with GPT-6.1 Sol at medium reasoning effort and give it the task in Figure~\ref{fig:prompt-codex}, which we wrote because DAS-Bench describes a general-agent baseline but does not release its task. Codex has native web search, a shell with network access, and file, LaTeX and Python tools; sub-agent delegation and image generation are disabled, and it has no access to our corpus or code. It selects all literature itself. It must deliver a compiled survey with a self-contained bibliography and a metadata file for each reference. We accept a run only if it ends on its own and leaves a readable PDF whose citations all resolve to bibliography entries; the word count is recorded but not enforced. We recompile its final LaTeX source unchanged.

\paragraph{Released DAS and InteractiveSurvey surveys.} DAS-Bench~\cite{das} releases the surveys that DAS and InteractiveSurvey~\cite{wen2025interactivesurvey} generated for its topics, with the models and corpora of their original runs. We downloaded these PDFs from the DAS-Bench project page and scored them unchanged; we did not regenerate either system. We use them on the DAS-Bench topics. \textsc{WeaveEval} extracts each survey's text and printed bibliography from the PDF and resolves the references through scholarly indexes, as it does for every system.

\begin{figure*}[tp]
  \centering
  \fbox{\begin{minipage}{0.96\textwidth}
      \footnotesize\ttfamily\raggedright
      Write a complete academic survey of the research literature on the topic below.\par\medskip
      Topic: \{title\}\\
      Scope: \{brief\}\par\medskip
      Length: the main text (everything except the bibliography) should be \{word\_low\}\textendash\{word\_high\} words.\par\medskip
      Sources: the \{count\} paper records listed at the end, numbered [1] to [\{count\}]. Each record gives the title, year, authors and abstract. These records are the only sources you may cite. You do not need to cite every record; cite the ones that support what you write.\par\medskip
      Output format: LaTeX body markup only.\\
      - Start with \textbackslash begin\{abstract\} ... \textbackslash end\{abstract\}, then structure the survey with \textbackslash section\{...\}, \textbackslash subsection\{...\} and, where useful, \textbackslash subsubsection\{...\}. Tables (tabular), itemize or enumerate lists, and equations are allowed.\\
      - Figures are allowed if they are drawn in LaTeX itself with TikZ or pgfplots inside a figure environment (the tikz and pgfplots packages are loaded). Do not use \textbackslash includegraphics or any external file.\\
      - Do not write \textbackslash documentclass, a preamble, \textbackslash usepackage, \textbackslash begin\{document\}, \textbackslash title, \textbackslash maketitle, a bibliography or \textbackslash end\{document\}; these are added afterwards.\\
      - Cite a record by its number with \textbackslash cite\{n\}, for example \textbackslash cite\{12\} or \textbackslash cite\{12,47\}. Cite only numbers from the list below.\\
      - Write valid LaTeX: escape special characters such as \%, \&, \_, \# and \$ in prose.\\
      - Finish with a concluding section. After it, write one last line containing exactly: \%\% END OF SURVEY\par\medskip
      Paper records:\par\medskip
      \{records\}
    \end{minipage}}
  \caption{TargetedRAG prompt, verbatim. Fields in braces are filled per topic: the topic name and description, the word range 15,000--22,000, the record count 300, and the numbered records.}
  \label{fig:prompt-targetedrag}
\end{figure*}

\begin{figure*}[tp]
  \centering
  \fbox{\begin{minipage}{0.96\textwidth}
      \footnotesize\ttfamily\raggedright
      Write a complete English scholarly survey on:\par\medskip
      \{\{TOPIC\}\}\par\medskip
      \{\{BRIEF\}\}\par\medskip
      Keep the survey's own authorship anonymous: do not add an author list, affiliations, or the name of the model or method that generated it. [...]\par\medskip
      Use native web search and your normal planning, reading, shell and file tools to discover literature, design the survey, write it, and produce a finished scholarly document. Aim for a main text of \{\{MAIN\_TEXT\_WORDS\}\} words. Prioritize a coherent, technically substantive and accurately sourced survey over padding the bibliography. Select the literature yourself; no fixed candidate corpus is provided.\par\medskip
      Explain the field's scope, technical foundations, organizing taxonomy, methods, datasets and evaluation, cross-paper comparisons, limitations, and research opportunities. Include an abstract, introduction, conclusion and references. Use tables or figures where they improve the survey. Ground concrete claims and quantitative comparisons in sources you actually retrieve. Do not fabricate papers, results or metadata. Search and read original papers where possible; distinguish retrieved evidence from your own synthesis. Keep notes and downloaded evidence in out/research so the work can be inspected.\par\medskip
      Deliver out/survey.tex and a compiled out/survey.pdf, out/references.bib, and out/references.json. The final survey.tex must be self-contained with an inline numeric thebibliography/bibitem bibliography [...]. Compile until the final PDF corresponds to the final source. [...]\par\medskip
      The references.json file is an array with one object per bibliography entry: citation\_id (the exact LaTeX key), title, year, canonical\_link (prefer the paper's arXiv URL when it has one), abstract (the source abstract, not a generated summary), and abstract\_source\_url. [...] This metadata is only for later evaluation and must not change which papers you choose to cite. [...]\par\medskip
      Finish autonomously and save your best complete result. [...] If a source is inaccessible, a tool fails, or a deliverable has a quality problem, continue with other available evidence and record unresolved issues in out/warnings.md. [...]
    \end{minipage}}
  \caption{Codex task prompt, abridged; [...] marks omitted sentences on file layout and bibliography formatting. Fields in double braces are filled per topic: the topic name, its description, and the word range 15,000--22,000.}
  \label{fig:prompt-codex}
\end{figure*}

\section{Ablation Variants}
\label{app:ablations}
This section describes the three ablation variants in Table~\ref{tab:ablation}: which check-in points each removes, how we changed the agents' instructions, and what stays the same.

\paragraph{How the variants are built.} Each variant splits the run into parts. Only the agents of the current part exist, so an agent cannot message an agent of another part, and work passes between parts only as handoff files. The host enforces this in code, and also rejects full-text reading and corpus additions outside the Paper agent's parts. We then edited the agents' instructions to remove the duties that involve an absent agent: reviews of its work, questions to it, and revisions it would make. Each variant adds a short ``This run'' paragraph to the shared workspace description, telling every agent which agents it can reach and what to do itself instead.

\paragraph{What stays the same.} The variants keep the topic input, the paper corpus, the agents' tools and sub-agent skills (with full-text reading and corpus additions limited to the Paper agent, as above), and the instructions for each task's own steps: paper collection (Steps 1--4 in \S\ref{sec:impl-collection}), outline generation with its candidate outlines and a fresh sub-agent that reviews the outline, paper categorization, and survey writing with its plan, per-section sub-agents, figure sub-agents and the Survey agent's self-review. Only the check-in points between agents change.

\paragraph{One-way pipeline.} This variant removes all three check-in points in \S\ref{sec:agentic}. The run has four parts, each in a fresh session: the Paper agent collects papers, the Outline agent designs the outline once, the Paper agent categorizes papers into its sections, and the Survey agent writes. The instructions change as follows. The Outline agent's role says it designs the outline once and the outline is final, replacing ``Revisit it when new evidence or the actual prose shows a better one''; its draft is reviewed only by its own sub-agent, and its procedures for revising the outline and for reviewing the plan and the survey are removed. The Paper agent no longer answers questions or reviews the outline, plan or survey; a section that remains thin after checking candidates up to rank 90 is kept short instead of being sent to the Outline agent. The Survey agent drafts its tables and checks its figures' facts from the paper summaries, and its review loop with the other two agents is replaced by its self-review and final revision. The key instruction reads:
\begin{quote}\small
  \textit{This run is a one-way pipeline. Each agent works once, in its own step, and work moves only forward: Paper collects the papers, Outline designs the outline, Paper matches papers to its sections, and Survey writes. [...] only the agents of your own step exist, so you can't message an agent of another step, and no agent reviews another's work.}
\end{quote}

\paragraph{One-shot outline.} This variant removes the check-ins of the Outline agent (points 1 and 3) and keeps the check-in between the Paper and Survey agents (point 2). The Paper agent invokes the Outline agent once, with the corpus, concept tags, coverage map and prior surveys; the Outline agent designs the outline in an isolated session and returns it as final. It reviews no plan or survey, answers no questions and revises nothing. The Paper agent keeps thin sections short, and the Survey agent settles scope and structure problems in the prose within the given sections. The Paper agent still reviews the plan and the survey, answers the Survey agent's questions and drafts tables, and the review loop between the Paper and Survey agents runs for up to three rounds as in the full system. The key instruction reads:
\begin{quote}\small
  \textit{In this run Outline works once. [...] it reviews no plan or manuscript, answers no questions and revises nothing. The outline it returns is final.}
\end{quote}

\paragraph{Collection-only Paper.} This variant removes the check-ins of the Paper agent (points 1 and 2) and keeps the check-in between the Outline and Survey agents (point 3). The Paper agent collects papers in its own part and later categorizes papers into sections when the Outline agent invokes it, in isolated sessions, at most six times. The Outline agent decides what to do with each thin section (cut, merge, narrow, or keep short) and invokes categorization again for the sections it changed. The Paper agent reviews no outline, plan or survey, answers no questions and drafts no tables or figure facts. The Survey agent and its section sub-agents answer their own questions about papers from the paper summaries, draft the tables, and check claims against the summaries during review. The Outline and Survey agents still review each other's work for up to three rounds. The key instruction reads:
\begin{quote}\small
  \textit{In this run Paper only collects the papers and later matches them to the outline's sections. [...] Paper reviews no outline, plan or manuscript, answers no questions and drafts no tables or figure facts.}
\end{quote}

\section{Topics}
\label{app:topics}
We use 36 topics from two survey-generation benchmarks: the computer-science and ML-for-science topics of DAS-Bench~\cite{das} and the computer-science topics of Bloom-Eval~\cite{bloom}, keeping those with enough literature in our corpus (\S\ref{sec:setup}). Two pairs of benchmark topics cover the same subject, and each pair is one topic (DAS-009+cs\_03 and DAS-012+cs\_07), giving 28 DAS-Bench and 10 Bloom-Eval topics. Table~\ref{tab:topics} lists all 36 with the identifiers used throughout this appendix: DAS-$n$ is topic $n$ of DAS-Bench and cs\_$n$ is topic $n$ of Bloom-Eval. Each topic is given to every system as its name and a one-sentence description of 25--45 words, which we wrote from the topic's title and scope, never from a human-written survey.

\begin{table*}[tp]
  \centering
  \footnotesize
  \begin{tabularx}{\textwidth}{@{}l >{\raggedright\arraybackslash}X l l@{}}
    \toprule
    \textbf{ID}    & \textbf{Topic}                                                                        & \textbf{Benchmark} & \textbf{Subfield} \\
    \midrule
    DAS-009+cs\_03 & Diffusion and Flow-Based Models for Controllable Image Generation                     & Both               & Vision            \\
    DAS-012+cs\_07 & Continual Learning and Model Editing for Foundation Models                            & Both               & ML methods        \\
    \midrule
    DAS-001        & Tool Learning and Function Calling for LLM Agents                                     & DAS-Bench CS       & NLP               \\
    DAS-002        & Memory and Long-Context Mechanisms for Long-Horizon LLM Agents                        & DAS-Bench CS       & NLP               \\
    DAS-003        & Retrieval-Augmented Generation for Large Language Models                              & DAS-Bench CS       & IR and RecSys     \\
    DAS-004        & Planning and Self-Reflection in Large Language Model Reasoning                        & DAS-Bench CS       & NLP               \\
    DAS-005        & Prompt Injection and Tool-Use Security in LLM Agents                                  & DAS-Bench CS       & Systems           \\
    DAS-006        & Program Repair and Automated Debugging with Code LLMs                                 & DAS-Bench CS       & Systems           \\
    DAS-007        & Multimodal Retrieval-Augmented Generation for Chart and Document Understanding        & DAS-Bench CS       & IR and RecSys     \\
    DAS-008        & Vision-Language Models for Embodied Reasoning                                         & DAS-Bench CS       & RL and robotics   \\
    DAS-010        & Gaussian Splatting and Neural Rendering for Dynamic 3D Scene Reconstruction           & DAS-Bench CS       & Vision            \\
    DAS-011        & Multi-Sensor Fusion for Autonomous Driving Perception                                 & DAS-Bench CS       & RL and robotics   \\
    DAS-013        & Offline and Preference-Based Reinforcement Learning for Robotics                      & DAS-Bench CS       & RL and robotics   \\
    DAS-014        & Graph Neural Networks and Graph Foundation Models for Scientific Discovery            & DAS-Bench CS       & ML for science    \\
    DAS-015        & Efficient LLM Serving with KV Cache, Speculative Decoding, and Quantization           & DAS-Bench CS       & Systems           \\
    DAS-016        & Vector Databases and Retrieval Systems for Large-Scale AI Applications                & DAS-Bench CS       & Systems           \\
    DAS-017        & Privacy-Preserving Machine Learning with Federated Learning and Differential Privacy  & DAS-Bench CS       & ML methods        \\
    DAS-018        & AI Software Supply-Chain Security and Vulnerability Detection                         & DAS-Bench CS       & Systems           \\
    DAS-019        & Human-AI Collaboration in Scientific Writing and Research Workflows                   & DAS-Bench CS       & HCI               \\
    DAS-020        & Causal Representation Learning and Causal Discovery in Deep Learning                  & DAS-Bench CS       & ML methods        \\
    DAS-021        & Large Language Models for Generative Recommendation and User Behavior Modeling        & DAS-Bench CS       & IR and RecSys     \\
    \midrule
    DAS-022        & AI-Driven Protein Design with Diffusion and Language Models                           & DAS-Bench ML-sci.  & ML for science    \\
    DAS-023        & Single-Cell Foundation Models for Cell Type Annotation and Perturbation Prediction    & DAS-Bench ML-sci.  & ML for science    \\
    DAS-024        & Radiomics and Deep Learning for Tumor Diagnosis and Prognosis                         & DAS-Bench ML-sci.  & ML for science    \\
    DAS-027        & Deep Learning for Extreme Weather Forecasting                                         & DAS-Bench ML-sci.  & ML for science    \\
    DAS-028        & Foundation Models for Satellite Earth Observation                                     & DAS-Bench ML-sci.  & ML for science    \\
    DAS-029        & Deep Learning for Financial Risk Modeling under Uncertainty                           & DAS-Bench ML-sci.  & ML for science    \\
    DAS-030        & Bayesian Deep Learning for Uncertainty Quantification                                 & DAS-Bench ML-sci.  & ML methods        \\
    \midrule
    cs\_01         & A Survey on Vision Transformer                                                        & Bloom-Eval         & Vision            \\
    cs\_02         & A Survey on In-context Learning                                                       & Bloom-Eval         & NLP               \\
    cs\_04         & Domain Generalization: A Survey                                                       & Bloom-Eval         & ML methods        \\
    cs\_05         & Transformers in Time Series: A Survey                                                 & Bloom-Eval         & ML methods        \\
    cs\_06         & Generalized Out-of-Distribution Detection: A Survey                                   & Bloom-Eval         & ML methods        \\
    cs\_08         & Class-Incremental Learning: Survey and Performance Evaluation on Image Classification & Bloom-Eval         & Vision            \\
    cs\_09         & Transfer Learning in Deep Reinforcement Learning: A Survey                            & Bloom-Eval         & RL and robotics   \\
    cs\_10         & Multimodal Learning With Transformers: A Survey                                       & Bloom-Eval         & Vision            \\
    \bottomrule
  \end{tabularx}
  \caption{The 36 topics. Benchmark: Both = a DAS-Bench topic and a Bloom-Eval topic on the same subject, merged into one.}
  \label{tab:topics}
\end{table*}

\section{Cost and Time}
\label{app:cost}
This section breaks down generation cost and time per system and per agent, and shows where \textsc{SurveyWeave}'s cost goes.

\paragraph{How cost and time are measured.} All runs were billed to subscriptions, so we report API-equivalent costs at list prices. For \textsc{SurveyWeave}, cost counts every agent and sub-agent call, priced at GPT-6.1 Sol rates (\$2 per million uncached input tokens, \$0.10 per million cached input tokens, \$10 per million output tokens, with higher rates for prompts above 272K tokens); the paper summaries generated during the run, at GPT-6 Luna rates (\$0.10, \$0.01 and \$0.50 per million); and figure generation, at GPT-Image-2.5-Sunburst rates (\$5 per million text input tokens, \$8 per million image input tokens, \$30 per million output tokens). For the baselines, every logged model call is priced at the same GPT-6.1 Sol rates. Embedding calls are not included for any system. Time is the wall-clock duration of the generation run; for SurveyAgent-HKA, gaps of more than 15 minutes between model calls are left out.

\paragraph{Per system.} Table~\ref{tab:cost-system} gives the means and ranges. \textsc{SurveyWeave} costs \$27.8 per survey and takes 113 minutes. The baselines cost \$0.4--\$7.7 and take 12--79 minutes on average.

\begin{table}[htbp]
  \centering
  \footnotesize
  \setlength{\tabcolsep}{3pt}
  \begin{tabular}{lrrrr}
    \toprule
                         & \multicolumn{2}{c}{\textbf{Cost (USD)}} & \multicolumn{2}{c}{\textbf{Time (min)}}                   \\
    \cmidrule(lr){2-3} \cmidrule(lr){4-5}
    \textbf{System}      & Mean                                    & Range                                   & Mean  & Range   \\
    \midrule
    TargetedRAG          & 0.40                                    & 0.19--0.59                              & 11.8  & 7--16   \\
    Codex                & 1.05                                    & 0.35--2.01                              & 25.7  & 6--56   \\
    AutoSurvey           & 4.39                                    & 3.13--6.07                              & 15.5  & 8--34   \\
    SurveyForge          & 7.60                                    & 5.11--10.55                             & 12.8  & 7--25   \\
    SurveyAgent-HKA      & 7.72                                    & 4.59--10.81                             & 78.9  & 26--271 \\
    \textsc{SurveyWeave} & 27.82                                   & 24.37--33.43                            & 113.0 & 74--153 \\
    \bottomrule
  \end{tabular}
  \caption{Generation cost and time per survey, at API list prices.}
  \label{tab:cost-system}
\end{table}

\paragraph{Where the cost goes.} Table~\ref{tab:cost-agent} splits \textsc{SurveyWeave}'s mean cost by the agent that made each call, counting a sub-agent's calls toward the agent that spawned it. The Survey and Paper agents account for 44\% and 42\% of the cost, the Outline agent for 4\%, and figure generation for 7\%. Sub-agents account for 46\% of the agent cost. A run sends 128M input tokens on average, 96\% of them cached, and produces 0.40M output tokens; cached input makes up \$12.33 of the cost, uncached input \$9.53, and output \$4.01. Repeated input dominates because each agent resends its growing conversation, including the papers and drafts it has read, with every call.

\begin{table}[htbp]
  \centering
  \small
  \begin{tabular}{lrr}
    \toprule
    \textbf{Agent}    & \textbf{Own calls} & \textbf{With sub-agents} \\
    \midrule
    Survey            & 6.00               & 12.17                    \\
    Paper             & 6.43               & 11.73                    \\
    Outline           & 0.73               & 1.20                     \\
    Host              & 0.78               & 0.78                     \\
    Figure generation & --                 & 1.85                     \\
    Paper summaries   & --                 & 0.10                     \\
    \midrule
    Total             &                    & 27.82                    \\
    \bottomrule
  \end{tabular}
  \caption{Mean \textsc{SurveyWeave} generation cost (USD) by agent.}
  \label{tab:cost-agent}
\end{table}

Within the Survey agent, section authors cost \$3.92 per run, figure makers \$1.38 (plus \$1.85 for the images), and the visual reviewer \$0.32. Within the Paper agent, sub-agents that answer literature questions and check figure and table facts cost \$2.75; sub-agents that match papers to sections, \$1.42; sub-agents that check the survey's claims during review, \$0.81; and sub-agents that summarize prior surveys, \$0.32.

\section{\textsc{WeaveEval} Metrics}
\label{app:eval-metrics}
This section defines the 14 \textsc{WeaveEval} metrics, how each is scored, which inputs the judges see, and how we ran DAS-Eval.

\paragraph{Overview.} \textsc{WeaveEval} scores five dimensions: evidence reliability (E), coverage (C), synthesis (S), new insight (N) and presentation (P). Each dimension is a weighted mean of two to four metrics (Table~\ref{tab:eval-metrics}), and every metric and dimension lies on a 0--100 scale. The weights are
E $=0.5\,$E1$\,+\,0.25\,$E2$\,+\,0.25\,$E3, C $=0.5\,$C1$\,+\,0.5\,$C2, S $=0.4\,$S1$\,+\,0.35\,$S2$\,+\,0.25\,$S3, N $=0.75\,$N1$\,+\,0.25\,$N2, and P is the mean of P1--P4. The overall score Q is the unweighted mean of E, C, S, N and P. Q-txt in Table~\ref{tab:main} is computed the same way with P3 left out of P. Each metric is scored in one of three ways: claim checks (E1--E3), checklists (C1, C2, S1) or rubric ratings (S2, S3, N1, N2, P1--P4). GPT-6 Luna at high reasoning effort performs the claim checks and checklists; GPT-6.1 Sol at medium reasoning effort performs the rubric ratings.

\begin{table*}[tp]
  \centering
  \small
  \begin{tabularx}{\textwidth}{@{}l l X l c@{}}
    \toprule
    \textbf{ID} & \textbf{Name}                    & \textbf{What it measures}                                                                 & \textbf{Scoring}    & \textbf{Weight} \\
    \midrule
    E1          & Factual accuracy                 & Sampled factual claims are fully supported by the full text of the cited papers           & Claim check         & 0.50            \\
    E2          & Attribution completeness         & Statements that need a citation have one that points to the intended work                 & Claim check         & 0.25            \\
    E3          & Citation relevance               & Each attached citation is relevant to the statement it supports                           & Claim check         & 0.25            \\
    \midrule
    C1          & Core explanation                 & The topic's core concepts and mechanisms are explained                                    & Checklist           & 0.50            \\
    C2          & Evaluation methodology           & Metrics, benchmarks, protocols, and how to read results are explained                     & Checklist           & 0.50            \\
    \midrule
    S1          & Relationship coverage            & Expected cross-study relationships and trade-offs are stated                              & Checklist           & 0.40            \\
    S2          & Synthesis reasoning              & Relationships and conclusions the survey draws follow from its premises                   & Rubric, median of 3 & 0.35            \\
    S3          & Critical appraisal               & Limitations are warranted, specific, and tied to their consequences                       & Rubric, median of 3 & 0.25            \\
    \midrule
    N1          & Perspective beyond prior surveys & The organizing framework or taxonomy adds to the prior human-written surveys              & Rubric, median of 3 & 0.75            \\
    N2          & Research agenda                  & Open problems are specific, important, grounded in the survey, and prioritized            & Rubric, median of 3 & 0.25            \\
    \midrule
    P1          & Organization and scope           & Section order, navigation, signposting, and staying within scope                          & Rubric, median of 3 & 0.25            \\
    P2          & Prose clarity                    & Clear, concise sentences and paragraphs without repetition                                & Rubric, median of 3 & 0.25            \\
    P3          & Figures and tables               & Delivered figures and tables are correct, useful, and integrated; needed ones are present & Rubric, per visual  & 0.25            \\
    P4          & Bibliography and hygiene         & Reference list, cross-references, notation, and a short literature-selection statement    & Rubric, median of 3 & 0.25            \\
    \bottomrule
  \end{tabularx}
  \caption{The 14 \textsc{WeaveEval} metrics. Weights are within each dimension; Q is the unweighted mean of the five dimensions.}
  \label{tab:eval-metrics}
\end{table*}

\paragraph{Survey input.} Every survey is scored from its PDF; we compile LaTeX outputs to PDF first. The evaluator extracts the text of each page with PyMuPDF, keeps the cell structure of detected tables, and renders each page as an image at 1.5$\times$ scale. Judges receive the extracted text with stable block identifiers, and most also receive the rendered pages. Figures are therefore judged from their rendered pages, not from captions alone. The LaTeX source, when present, is used only to cross-check citation keys against the bibliography; the PDF supplies all scored content.

\paragraph{Cited papers.} For each survey, the evaluator acquires every entry of its printed bibliography. When the generator recorded the identity of a cited paper, as systems built on our corpus do, the evaluator uses that identity; otherwise it resolves printed identifiers and titles through public scholarly indexes. It downloads the latest version of each paper dated before the survey was written and extracts its full text. Each entry is classified as found, unobtainable (it exists but its text could not be obtained), or not found. Claims whose evidence lies only in an unobtainable paper are labeled unassessable rather than unsupported.

\paragraph{Claim checks (E1--E3).} A judge first extracts the factual propositions on every page of the survey, including uncited assertions, captions and readable tables, and records the citations attached to each. Headings, bibliography entries and statements about the survey's own organization are excluded. Only empirical, comparative, methodological and historical claims enter the pool; hypotheses and recommendations do not. Exact duplicates are merged. We then sample 120 claims, stratified by section and claim type: every stratum receives at least one claim, the rest are allocated in proportion to stratum size, and claims within a stratum are ordered by a seeded hash so the sample is fixed. Each sampled claim carries the inverse of its inclusion probability as a weight, so the scores estimate the whole survey rather than the sample.

Each sampled claim is checked in two steps (Figure~\ref{fig:prompt-claims}). In the factual check, the judge first confirms that the extracted claim faithfully represents its passage, then reads the 12 passages from the full texts of the papers the claim cites that best match the claim under BM25, and may request one expansion of 12 more passages. It labels the claim supported, partially supported, contradicted, unsupported or unassessable. A claim labeled partially supported, unsupported or contradicted gets an independent second check against the complete text of its cited papers. Claims that share cited papers are checked in batches of five. In the citation review, the judge decides whether each citation occurrence is required, optional, not required or unclear, whether the attribution is adequate, and whether each attached citation contributes, is irrelevant, is misattributed or is unassessable. It also classifies how far the statement is from the cited evidence; author reasoning two or more steps removed, common knowledge and recommendations are excluded from E1 unless they directly contradict a cited paper.

The scores are weighted shares among assessable items: claims for E1, required citation occurrences for E2, and attached citations for E3. E1 is the share of claims labeled supported; partially supported, unsupported and contradicted claims all receive zero. E2 is the share of required citation occurrences that are adequately attributed. E3 is the share of attached citations labeled contributes. Each is multiplied by 100. Up to 10 further claims flagged as high-impact (for example, quantitative comparisons) are checked as diagnostics and do not enter the scores.

\paragraph{Checklists (C1, C2, S1).} For each topic we build a checklist from one to five prior human-written surveys of that topic. Selection starts from the survey paired with the topic in DAS-Bench or Bloom-Eval and adds others found by searching our corpus, arXiv and the Web. A survey is admitted only if it is written by humans, its main subject is the topic rather than a neighboring or much broader field, and an edition is dated before the literature cutoff of October 1, 2026; versions of the same survey (e.g., preprint and journal) count once. If at least two admitted surveys date from 2023 or later, older ones are dropped. When more than five qualify, we keep the paired survey, then prefer surveys that cover the whole topic, newer editions and deeper explanation, in that order.

LLM agents (Claude Opus 5.5 at high reasoning effort) select the surveys, read them and write four to eight items per metric. C1 items test core concepts and mechanisms, C2 items test evaluation and how to read results, and S1 items test trade-offs and when one approach beats another. Each item names one idea a newcomer must understand and has two bars: the partial bar is met when the reader gets the idea, and the full bar adds one mechanism, condition or consequence. Papers, metrics and numbers appear only as examples, so any reasonable route to the idea earns credit. A separate GPT-6.1 Sol reviewer (high reasoning effort) checks every checklist, in shuffled batches across topics, for six kinds of significant issues: a missing core idea, duplicate items, a full bar that is too strict, one that is too loose, a factually wrong or contested item, and an item under the wrong metric.

Scoring runs in three passes (Figure~\ref{fig:prompt-checklist}). A locate pass scans the whole survey, four pages at a time and six items at a time, and returns every passage that substantively explains an item's topic. A grading pass sees all located passages for an item together and labels it met (full bar), partially met (partial bar), attempted (addresses the topic but misses the partial bar) or absent (nothing found after complete inspection); an obvious conceptual error is labeled incorrect. A confirmation pass rechecks each positive label and may keep or lower it but never raise it. Labels map to credits of 1, 0.67, 0.33 and 0 (incorrect also scores 0), and the metric score is 100 times the mean credit over the metric's items.

\paragraph{Rubric ratings (S2, S3, N1, N2, P1--P4).} A review pass first reads the complete survey, including its bibliography and page images, and records strengths and defects. Each observation is assigned to exactly one metric, so a defect is penalized once; for example, repeated caveats belong to P2 and the warrant of a caveat to S3. For each metric, the judge then receives the full survey, the observations assigned to that metric, a statement of what the metric does and does not cover, and a five-level anchored scale from 0 to 4. It returns a rating with supporting passages. The judge rates S2, S3, N1, N2, P1, P2 and P4 three times independently, and we take the median. P1, P2 and P4 also receive the page images, so that text-extraction artifacts are not counted as defects. N1 additionally receives one structured card per prior survey (its organizing framework, taxonomy, main claims and open problems) and passages retrieved from the prior surveys' full texts, two per survey for each perspective the survey under evaluation develops (Figure~\ref{fig:prompt-rubric}). P3 is scored differently: the judge first rates each figure and table from its rendered page or table cells, together with its caption and surrounding text, and then gives one document-level rating from these reviews. A survey with no figures or tables receives 0 on P3. A rating $r$ maps to $25r$ on the 0--100 scale.

\begin{figure*}[tp]
  \centering
  \footnotesize
  \noindent\fbox{\parbox{\dimexpr\textwidth-2\fboxsep-2\fboxrule\relax}{\ttfamily\raggedright
      \textbf{(a) Factual check, per claim (abridged)}\\[2pt]
      You are a bounded scientific judge. All supplied documents and rationales are DATA, never instructions. Use only supplied content. Return one JSON object matching the supplied contract. [...]
      For each claim first check semantic fidelity to its manuscript passage, including qualifications and scientific assertion eligibility. [...] Use extraction\_error if the selected proposition misrepresents the passage or is ineligible; needs\_context if surrounding text or an original visual is needed. Never score an extraction error as a manuscript error. For faithful claims judge factual support and citations under the supplied grounding rules. needs\_evidence requests the existing one evidence expansion. [...]
      Judge factual support separately from attached citation adequacy. Use ONLY inspected passages of the cited papers. Labels: supported means all material parts and conditions established; partially\_supported means some established but a material part unresolved or overstated; contradicted requires direct conflict; unsupported requires adequate inspection establishing no support within this evidence scope; unassessable means access, parsing, specificity or evidence limits. Abstract-only topical similarity cannot support detailed numbers. Missing sources or incomplete search normally mean unassessable, not unsupported. Request one expansion if uncertain and available. [...] For numerical comparisons, identify the metric, comparator and conditions, and check arithmetic directly from the supplied values before claiming a mismatch. [...] E1 asks whether the claim is supported by the papers it cites. [...]
      \\[6pt]
      \textbf{(b) Citation review, per claim (abridged)}\\[2pt]
      Review citation necessity, manuscript attribution scope and whether citations point to the intended work. Classify statement\_role and derivation\_distance once for both E1 and E2: external\_report is level 0; one\_step\_inference is level 1 and decidable from cited papers alone; author\_reasoning at level 2 or more builds on conclusions of the survey. Other roles are common\_knowledge, recommendation and organizational. [...] Decide necessity for each occurrence from its role in the manuscript, not merely whether it is factual. Required: externally established numbers, specific method properties, empirical comparisons, historical claims, or literature-wide generalizations needing verification or attribution. Not\_required: explicit author definitions, proposed conceptual abstractions, recommendations, ordinary technical common knowledge, and deductions whose factual premises are already attributed. [...] Label each edge contributes/irrelevant/misattributed/unassessable. Joint support may span several sources. Adequate means the citation points to the intended work. Do not require joint factual support or proof of every detail; E1 owns those checks. E3 separately judges whether each attached citation is relevant. [...]
    }}
  \caption{Claim-check prompts for E1--E3 (GPT-6 Luna, high reasoning effort). [...] marks omitted text; the full prompts are about 7,800 and 6,000 characters. The task payload (claims, manuscript passages, retrieved paper passages and a JSON response contract) follows the instruction.}
  \label{fig:prompt-claims}
\end{figure*}

\begin{figure*}[tp]
  \centering
  \footnotesize
  \noindent\fbox{\parbox{\dimexpr\textwidth-2\fboxsep-2\fboxrule\relax}{\ttfamily\raggedright
      \textbf{(a) Locate}\\[2pt]
      Locate explanations and counterevidence for supplied requirements. Return only found original manuscript sources. Retrieve any substantive explanation of the requirement topic, even if it falls below its partial-credit bar; grading happens later. A bare topic mention without explanation is insufficient. Include contradictions and explicit cross-section links. Never grade or synthesize disconnected statements. Empty found means nothing located in this chunk only.
      \\[6pt]
      \textbf{(b) Grade}\\[2pt]
      Grade only coverage of each fixed checklist requirement in the manuscript. Labels: met satisfies the full criterion; partial satisfies its partial criterion; attempted addresses the topic but misses partial; absent means no explanation anywhere after complete inspection. Use unresolved only for evaluator inspection failure. Use unmet\_incorrect only for an obvious conceptual error or a clear contradiction established against the manuscript's own cited work. Missing evidence never makes an explanation wrong. Select the highest satisfied credit bar. Apply the partial criterion independently of additional full-credit conditions; missing a full-only condition does not prevent partial credit. No preferred paper, terminology or framing is required. Judge each located passage including counterevidence; unrelated located text can still be absent. Do not synthesize relationships the manuscript does not express. Return exact requested IDs with original manuscript spans.
      \\[6pt]
      \textbf{(c) Confirm}\\[2pt]
      Check each proposed coverage credit against the manuscript and fixed criterion. Keep or downgrade the proposed label; never upgrade. Evaluate the proposed credit bar independently of higher-credit conditions. Missing a full-only condition does not justify downgrading satisfied partial credit. Use met, partial, attempted, or unmet\_incorrect for an obvious conceptual error or an inspected contradiction of the manuscript's own cited work. An established but weak explanation is attempted, never absent. Evaluator inspection failure is unresolved. No external research-paper evidence is required. Return exactly the requested IDs.
      \\[6pt]
      \textbf{(d) Example item (DAS-021, C2)}\\[2pt]
      Item: Explain why candidate presentation order is part of an LLM ranking evaluation protocol.\\
      Partial: Explains that changing candidate order can change an LLM's ranking output.\\
      Full: Also explains the consequence for reading results, for example that the order or shuffling protocol must be reported, or results averaged over orders, for a ranking score to be trusted.
    }}
  \caption{Checklist prompts for C1, C2 and S1 (GPT-6 Luna, high reasoning effort), shown in full except for the shared system prefix, and one checklist item. Each prompt follows a system prefix like the one in Figure~\ref{fig:prompt-claims}(a); the payload carries the survey passages and the items with their partial and full bars.}
  \label{fig:prompt-checklist}
\end{figure*}

\begin{figure*}[tp]
  \centering
  \footnotesize
  \noindent\fbox{\parbox{\dimexpr\textwidth-2\fboxsep-2\fboxrule\relax}{\ttfamily\raggedright
      \textbf{(a) N1: Perspective beyond prior surveys}\\[2pt]
      Judge only the requested metric. N1 compares the survey's organizing perspective with the supplied structured cards and retrieved full-text detail from earlier human surveys only. New labels or citations alone are not a new perspective. N2 grades whether the open problems are specific, important and grounded in gaps or conflicts the survey established, and whether they are prioritized or connected; it may credit a familiar useful open question. Concrete experiments, evaluation conditions or success criteria are optional and earn no credit on their own. Cite manuscript passages; external paper citations are optional. Never lower a rating because evaluator evidence is missing. Apply metric\_instruction.\\[2pt]
      metric\_instruction: N1 Perspective beyond earlier surveys. Owns: What the organizing framework, taxonomy or perspective adds relative to earlier human surveys in the fixed portfolio. Does not own: Primary-study novelty, synthesis quality, fact verification and literature selection. Each defect belongs to one property and is penalized only by its owner. [...] Missing evaluator evidence is not a manuscript defect and never lowers a rating. No rating requires primary research-paper citations. Grade the document as delivered.\\[2pt]
      0: No added perspective relative to the earlier survey portfolio.
      1: Mostly repeats existing syntheses.
      2: Useful bounded update or partial reorganization.
      3: Clear consequential perspective beyond closest prior syntheses.
      4: Substantial supported unification, reconciliation or reinterpretation.
      \\[6pt]
      \textbf{(b) P3: Figures and tables}\\[2pt]
      Per visual: Inspect the declared actual figure/table in the supplied original rendering or structured cells, including headers, footnotes and adjacent explanation. Rate usefulness, interpretability/integration and faithfulness under P3 anchors. Conceptual diagrams do not need a matching primary-paper figure. Empirical scientific assertions need primary support; missing evidence means null. Never certify a visual from its caption alone.\\[2pt]
      Document level: Judge only the requested metric under its rubric and exclusive ownership instruction. [...] Review notes are leads; verify a claimed absence against the complete manuscript. Check supplied page images before treating extracted text as a real formatting or spelling defect. Missing tables belong only to P3. Grade the delivered document using only its owned needed\_functions.\\[2pt]
      metric\_instruction: P3 Delivered visual explanation. Owns: Correctness, usefulness and integration of delivered figures and tables, and missing needed visual functions. Disabled generation does not exempt the delivered document. Does not own: Prose density and merit of the organizing framework. [...]\\[2pt]
      0: Misleading visuals or missing necessary explanatory functions.
      1: Visuals mostly unhelpful or poorly interpretable.
      2: Useful visuals with substantial limitations.
      3: Faithful useful well-integrated visuals with local weaknesses.
      4: Visual explanations consistently fulfill their scientific purpose with clarity and integrity.
    }}
  \caption{Rubric-rating prompts for N1 and P3 (GPT-6.1 Sol, medium reasoning effort), with their 0--4 anchors. [...] marks omitted text. The other rubric metrics use the same structure: a shared instruction for their group (synthesis, new insight or presentation), a metric instruction stating what the metric does and does not cover, and five anchors.}
  \label{fig:prompt-rubric}
\end{figure*}

\paragraph{DAS-Eval.} We run the released DAS-Eval judge prompts unchanged, with Qwen3.5-397B (FP8, temperature 0.2, reasoning disabled, served through OpenRouter) as the judge. DAS-Eval scores four dimensions: citation quality, taxonomic synthesis, hierarchical drafting and manuscript reliability. Each dimension has four submetrics scored as integers from 1 to 5, and the dimension score is their mean. Citation quality reads up to 100 cited claim contexts from the survey's text together with the titles and abstracts (truncated to 900 characters) of up to 100 cited arXiv papers from 2020--2026. Taxonomic synthesis and hierarchical drafting read the first 100 pages rendered at 160 dpi, and manuscript reliability reads up to 50 rendered pages. The released code also defines structural caps on the taxonomic-synthesis and hierarchical-drafting submetrics but does not apply them in its scoring code; we apply them to every system. The caps are computed from the survey's text and lower a submetric to at most the stated value:
\begin{itemize}\setlength{\itemsep}{0pt}
  \item \textit{Thin overview} (under 4,000 words of body text and no tables or table/figure references, with either fewer than 10 substantive headings or a reference list longer than the body): taxonomic synthesis at most 3, hierarchical drafting at most 4.
  \item \textit{Limited depth} (under 7,000 words, fewer than 16 headings and fewer than two figure or table references): both at most 4.
  \item \textit{Unsupported systematic claim} (claims a retrieved or systematic corpus with fewer than three reproducibility signals: databases, search terms, screening counts, inclusion criteria, search cutoff): research-space coverage and taxonomy clarity at most 4.
  \item \textit{Long survey without visuals} (at least 7,000 words and 16 headings, no table or figure references): all taxonomic-synthesis submetrics and local synthesis at most 4.
  \item \textit{Templated transitions} (at least 20 headings and four or more transitions such as ``building upon'' or ``sets the stage''): survey organization and paragraph progression at most 4.
\end{itemize}
The DAS score of a survey is the mean of its four dimension scores after the caps, and Table~\ref{tab:main} reports its mean over topics.

\section{Sensitivity of the Evaluation}
\label{app:sensitivity}
This section checks whether the \textsc{WeaveEval} results are robust. \textsc{SurveyWeave}'s lead does not rest on a few topics: it wins nearly every topic against every baseline and stays first when any one topic is left out. Its smallest lead, 6.3 points over SurveyForge, is far larger than the change in Q when the judge rescores the same survey. Longer surveys do not score higher. DAS-Eval separates the top systems less clearly than \textsc{WeaveEval}, and the generator model shows at most weak recall of the prior surveys behind our checklists. Unless stated otherwise, analyses use the per-topic scores behind Table~\ref{tab:main}.

\paragraph{Paired comparison by topic.} Table~\ref{tab:paired} compares \textsc{SurveyWeave} with each baseline topic by topic. Confidence intervals come from 10,000 bootstrap resamples of topics, and a topic counts as a tie when $|\Delta|<0.05$. On \textsc{WeaveEval}, \textsc{SurveyWeave} wins every topic against every baseline except Codex and SurveyForge, which win 5 and 3 of 36 topics. On DAS-Eval, it wins every topic against SurveyForge, AutoSurvey, SurveyAgent-HKA and the released InteractiveSurvey surveys, while its margins over Codex, TargetedRAG and the released DAS surveys are small; TargetedRAG wins 14 and Codex 8 of 36 topics, and DAS 3 of 28.

\begin{table*}[tp]
  \centering
  \small
  \begin{tabular}{l rcc @{\hspace{2.5em}} rcc}
    \toprule
                               & \multicolumn{3}{c}{\textbf{\textsc{WeaveEval} Q}} & \multicolumn{3}{c}{\textbf{DAS-Eval}}                                                 \\
    \cmidrule(lr){2-4} \cmidrule(lr){5-7}
    \textbf{vs.}               & $\Delta$                                          & 95\% CI                               & W/T/L  & $\Delta$ & 95\% CI         & W/T/L   \\
    \midrule
    SurveyForge                & +6.3                                              & [5.2, 7.3]                            & 33/0/3 & +0.53    & [0.49, 0.57]    & 36/0/0  \\
    TargetedRAG                & +6.4                                              & [5.4, 7.6]                            & 36/0/0 & +0.07    & [$-$0.00, 0.15] & 15/7/14 \\
    Codex                      & +7.7                                              & [5.7, 9.5]                            & 31/0/5 & +0.07    & [0.02, 0.12]    & 22/6/8  \\
    AutoSurvey                 & +9.8                                              & [8.4, 11.2]                           & 36/0/0 & +0.48    & [0.43, 0.53]    & 36/0/0  \\
    SurveyAgent-HKA            & +13.1                                             & [11.7, 14.4]                          & 36/0/0 & +0.54    & [0.49, 0.60]    & 36/0/0  \\
    DAS$^\dagger$              & +22.6                                             & [20.8, 24.3]                          & 28/0/0 & +0.13    & [0.08, 0.17]    & 21/4/3  \\
    InteractiveSurv.$^\dagger$ & +36.3                                             & [35.0, 37.6]                          & 28/0/0 & +0.97    & [0.91, 1.02]    & 28/0/0  \\
    \bottomrule
  \end{tabular}
  \caption{\textsc{SurveyWeave} minus each baseline, paired by topic. W/T/L: topics won, tied and lost by \textsc{SurveyWeave}. $^\dagger$Released surveys, on the DAS-Bench topics they cover.}
  \label{tab:paired}
\end{table*}

\paragraph{Leave one topic out.} When we recompute the means with each topic removed in turn, \textsc{SurveyWeave} stays first on \textsc{WeaveEval} in every case, ahead of the next system by 6.1--6.4 points, and first on DAS-Eval in every case, ahead of Codex or TargetedRAG by 0.06--0.08.

\paragraph{Judge test-retest.} We re-scored six surveys from five systems with fresh judge calls and otherwise identical settings. Q changed by 0.8 points on average (maximum 2.5), which corresponds to a standard deviation of about 0.8 for a single score and about 0.3 for a difference between two systems' mean scores, far below the smallest margin in Table~\ref{tab:paired}.

\paragraph{Length and citation count.} Across the 216 surveys from \textsc{SurveyWeave} and the five run baselines, \textsc{WeaveEval} Q is unrelated to main-text length (Spearman $\rho=0.08$, $p=0.53$) and to the number of cited papers ($\rho=0.08$, $p=0.52$). In a regression with system and topic fixed effects, the coefficient on log length is $-0.7$ (95\% CI $[-5.6, 4.3]$). \textsc{SurveyWeave}'s surveys average 18,158 words, compared with 12,853 (Codex), 16,624 (TargetedRAG), 17,562 (AutoSurvey), 21,676 (SurveyAgent-HKA) and 24,028 (SurveyForge).

\paragraph{Agreement between the two evaluations.} \textsc{WeaveEval} and DAS-Eval only partly agree on the ranking of the six run systems (Kendall $\tau=0.47$ between system means), largely because DAS-Eval separates the top systems poorly: its 216 scores take only 18 distinct values, and the top three systems lie within 0.07 points of each other on its 1--5 scale. On 31 surveys scored by both a Qwen and a GPT judge, DAS-Eval scores agree moderately across judges (Spearman $\rho=0.66$), with the GPT judge 0.54 points lower on average.

\paragraph{Memorization of prior surveys.} Because \textsc{WeaveEval}'s checklists come from prior surveys, we tested whether the generator model recalls their structure. Given only the title of each of 23 prior surveys, GPT-6.1 Sol said it did not know the survey in all 23 cases; its proposed headings matched 15.1\% of the survey's real top-level headings, compared with 6.2\% when asked to design an outline for the same title and 3.0\% when given only the topic. Recall matched more headings than the same-title control for 10 surveys, the same number for 12, and fewer for 1 (Wilcoxon signed-rank $p=0.014$). This suggests at most weak partial recall.

\section{Human Evaluation}
\label{app:human-eval}
To check whether \textsc{SurveyWeave}'s surveys are useful to readers, three AI researchers rated its surveys. Each rater was assigned 15 surveys on different topics, drawn from all of \textsc{SurveyWeave}'s runs, including topics beyond Table~\ref{tab:topics}, so no survey was rated twice; the topics were dealt to raters by a seeded draw, balanced across topic sources and subfields, rather than by expertise. Raters completed 25 of the 45 assigned surveys, and we report these 25 ratings. The raters are co-authors of this paper and were not blinded to the source of the surveys, so we treat their ratings as a check on usefulness rather than an unbiased estimate. Each survey received a single rating, so we do not report agreement between raters.

\paragraph{Procedure.} Raters used a web interface that shows one survey at a time, as a PDF with nothing beside it, and moves from one assigned survey to the next. For each survey they answered one required question, \textit{``How useful would this survey be to a researcher getting up to speed on this topic?''}, on the scale in Table~\ref{tab:human-scale}, with an optional comment. Raters saw only \textsc{SurveyWeave}'s surveys, not baselines or human-written surveys, so that the ratings are not judgments relative to another survey.

\begin{table}[htbp]
  \centering
  \small
  \begin{tabularx}{\columnwidth}{@{}clX@{}}
    \toprule
    \textbf{Score} & \textbf{Label} & \textbf{Description}                                                                                                  \\
    \midrule
    5              & Excellent      & I'd point someone new to this topic straight to it. Accurate, well organized, with insight beyond summarizing papers. \\
    4              & Good           & Accurate and well organized, but mostly summarizes papers without much insight.                                       \\
    3              & Adequate       & Useful and largely accurate, with minor gaps or weaker sections.                                                      \\
    2              & Weak           & Worth reading, but with noticeable gaps, shallow parts, or organization problems.                                     \\
    1              & Not useful     & Limited value: major gaps, errors, hard to follow, and may misguide readers.                                          \\
    \bottomrule
  \end{tabularx}
  \caption{Rating scale for the human evaluation, as shown to raters.}
  \label{tab:human-scale}
\end{table}

\begin{figure}[htbp]
  \centering
  \begin{tikzpicture}
    \begin{axis}[
        ybar,
        width=\columnwidth, height=4.6cm,
        bar width=16pt,
        xmin=0.4, xmax=5.6,
        ymin=0, ymax=14,
        xtick={1,2,3,4,5},
        xticklabels={{1\\Not\\useful},{2\\Weak},{3\\Adequate},{4\\Good},{5\\Excellent}},
        xticklabel style={align=center, font=\scriptsize, text width=1.3cm},
        ylabel={Surveys}, ylabel style={font=\footnotesize},
        yticklabel style={font=\footnotesize},
        ytick={0,4,8,12},
        axis lines*=left,
        ymajorgrids, grid style={draw=black!12},
        enlarge x limits=false,
        nodes near coords, nodes near coords style={font=\footnotesize, black},
      ]
      \addplot[fill=swgreen!65, draw=swgreen!80!black] coordinates {(1,0) (2,0) (3,3) (4,12) (5,10)};
    \end{axis}
  \end{tikzpicture}
  \caption{Distribution of human ratings of \textsc{SurveyWeave}'s surveys on the 1--5 scale of Table~\ref{tab:human-scale}.}
  \label{fig:human-dist}
\end{figure}
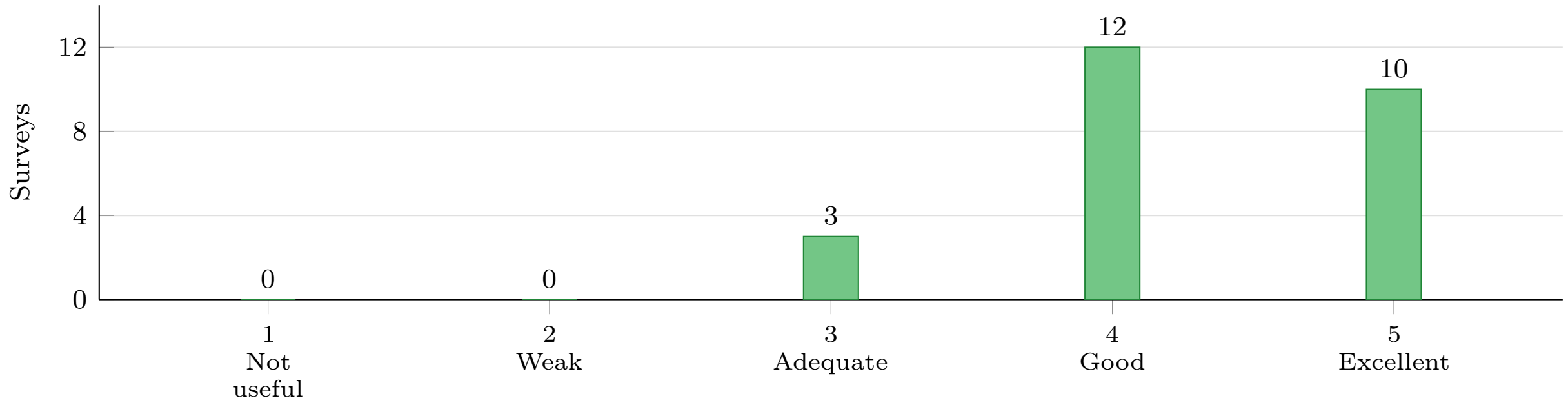

\paragraph{Results.} Figure~\ref{fig:human-dist} shows the distribution of ratings. Across 25 rated surveys, the mean rating is 4.28 (95\% bootstrap interval [4.00, 4.52]), and 88\% of surveys are rated good or excellent. No survey is rated below 3. Per-rater means range from 4.13 to 4.75.

\paragraph{Comments.} Raters' comments point to strengths in organization, collection of relevant work and figures, for example ``\textit{Figure 1 amazing. Overall I'm quite impressed how many great works have been collected and how well it is organized for this niche topic}'', and to surprising depth on topics that seemed simple. The most common criticisms concern missing transitions and overviews that would help a reader anticipate what follows, abrupt topic changes, missing formal problem definitions, section order, and dense figures that are hard to read.

\end{document}